\documentclass[lettersize,journal]{IEEEtran}

\usepackage{amsfonts}
\usepackage{mathrsfs}
\usepackage{amssymb}
\usepackage{wrapfig}
\usepackage{amsmath}
\usepackage{accents}
\usepackage{dsfont}
\usepackage{amsbsy}
\usepackage{setspace}
\usepackage{graphicx}
\usepackage{subfig}
\usepackage{url}
\usepackage{color}
\usepackage{epstopdf}
\usepackage{amssymb}
\usepackage{framed}
\usepackage{multirow}
\usepackage{mathtools}  
\usepackage{algorithm}
\usepackage{tabularx}
\usepackage{diagbox}
\usepackage{geometry}

\title{On the Regularity and Fairness of Combinatorial Multi-Armed Bandit}

\newtheorem{remark}{Remark}

\newcommand{\bE}{\mathds{E}}
\newcommand{\mc}{\mathcal}
\newcommand{\mb}{\mathbf}
\newcommand{\bs}{\boldsymbol}
\newcommand{\ol}{\overline}
\newcommand{\wt}{\widetilde}
\newcommand{\wh}{\widehat}
\newcommand{\id}{\mathds{1}}

\def\argmax{\operatornamewithlimits{arg\,max}}

\author{Xiaoyi Wu,~\IEEEmembership{Graduate Student Member,~IEEE} and Bin Li,~\IEEEmembership{Senior Member,~IEEE}}

\begin{document}
\maketitle
\newtheorem{theorem}{Theorem}
\newtheorem{lemma}{Lemma}
\newtheorem{claim}{Claim}
\newtheorem{proposition}{Proposition}
\newtheorem{corollary}{Corollary}
\newtheorem{definition}{Definition}
\newtheorem{assumption}{Assumption}
\newtheorem{remarks}{Remarks}

\begin{abstract}
The combinatorial multi-armed bandit model is designed to maximize cumulative rewards in the presence of uncertainty by activating a subset of arms in each round. This paper is inspired by two critical applications in wireless networks, where it's not only essential to maximize cumulative rewards but also to guarantee fairness among arms (i.e., the minimum average reward required by each arm) and ensure reward regularity (i.e., how often each arm receives the reward). In this paper, we propose a parameterized regular and fair learning algorithm to achieve these three objectives. In particular, the proposed algorithm linearly combines virtual queue-lengths (tracking the fairness violations), Time-Since-Last-Reward (TSLR) metrics, and Upper Confidence Bound (UCB) estimates in its weight measure. Here, TSLR is similar to age-of-information and measures the elapsed number of rounds since the last time an arm received a reward, capturing the reward regularity performance, and UCB estimates are utilized to balance the tradeoff between exploration and exploitation in online learning. By exploring a key relationship between virtual queue-lengths and TSLR metrics and utilizing several non-trivial Lyapunov functions, we analytically characterize zero cumulative fairness violation, reward regularity, and cumulative regret performance under our proposed algorithm. These theoretical outcomes are verified by simulations based on two real-world datasets.
\end{abstract}

\begin{IEEEkeywords}
Combinatorial Multi-Armed Bandit, Fairness, Service Regularity.
\end{IEEEkeywords}

\section{Introduction}







Combinatorial multi-armed bandit (CMAB) is a type of multi-armed bandit problem that involves choosing a subset of arms to be pulled simultaneously in each round. Once arms are pulled, each pulled arm will return a random reward assumed to be independently and identically distributed over rounds. The goal of CMAB is to maximize the cumulative rewards obtained from pulling selected arms with the unknown distribution of reward. The CMAB problems occur in many real-world network applications such as resource allocation (e.g. \cite{ortiz2019cbmos}), network routing (e.g. \cite{fu2022optimal}), and wireless user scheduling (e.g., \cite{gai2010learning,alipour2022multiuser,kang2020low}).


However, traditional CMAB problem formulation is not well adapted to some emerging applications, which have a high demand for fairness (i.e., guaranteeing minimum average reward required by each arm) and reward regularity (i.e., how often each arm receives a reward). For example, consider the problem of delivering interactive and panoramic scenes (e.g., panoramic video streaming and virtual reality) from an access point (AP) to multiple users. In this scenario, we need to maximize the average rate of users successfully viewing the desired content, which is unknown a priori and must be learned over time (see \cite{chen2021motion,chen2020thompson}). Meanwhile, in order to provide users with more satisfactory service, fairness among multiple users and a seamless viewing experience are supposed to be taken into account as well. Moreover, in the problem of scheduling multiple sensing sources to transmit time-sensitive information over unreliable wireless channels with unknown successful transmission probabilities,  a subset of sensing sources can transmit data simultaneously to one AP \cite{li2021efficient}. To keep the completeness and freshness of sensing information collected at AP, not only the system throughput but also fairness among sensing sources and the age of information from each sensing source should be considered. Please see the detailed discussions of these two motivating applications in Section \ref{sec:motivating}. Therefore, from the above two examples, we can see that the traditional CMAB framework cannot be applied to applications that require both fairness and reward regularity in addition to maximizing cumulative rewards. 


Recent studies demonstrate the growing interest in CMAB problems with fairness constraints (e.g., \cite{li2019combinatorial,patil2020achieving,liu2021efficient,liu2022combinatorial,bernasconi2022safe}). In particular, authors in \cite{li2019combinatorial} introduced virtual queues to track the fairness violation and incorporated them into the algorithm design together with the Upper Confidence Bound (UCB) weight \cite{auer2002finite} for estimating the rewards. They characterized cumulative regret and long-term fairness performance. Here, cumulative regret is defined to be the difference between the total reward obtained by some algorithm and the maximum possible total reward that could have been obtained by pulling a subset of arms with the largest mean rewards throughout the entire rounds.  Recently, the authors in \cite{liu2021efficient} proposed a pessimistic-optimistic algorithm that achieves state-of-the-art regret performance and a zero fairness constraint violation by properly selecting algorithmic parameters. Subsequently, many studies utilized the CMAB framework with fairness constraints in different applications, e.g.,  client selection in federated learning (e.g., \cite{xia2020multi, huang2020efficiency, zhu2022online}), crowdsourcing (e.g., \cite{gao2021budgeted, li2022fair}), and multi-agent scenarios (e.g., \cite{verma2020stochastic,yang2021cooperative}). However, none of these works address the reward regularity performance. After introducing the regularity metrics, solving the problem becomes even more difficult due to the existence of a strong coupling between fairness constraints and reward regularity.

The reward regularity is similar to service regularity (e.g., \cite{li2014throughput,li2015wireless}) or the age of information (AoI) (e.g., \cite{lu2018age,kadota2018optimizing,kadota2019minimizing} and see \cite{sun2019age} and \cite[Ch. 8] {pappas2022age} for an overview) in networking areas. These works characterized the steady-state service regularity performance or average AoI performance. However, they did not address the algorithm design in an unknown environment, where the efficient algorithm design not only makes decisions but also learns the unknown system statistics. Recent work \cite{li2021efficient} integrated AoI metrics and UCB estimates into the algorithm design and revealed the tradeoff between the running average AoI (reward regularity in our context) and cumulative regret performance. Subsequently, some studies leveraged the framework of CMAB to solve optimization problems involving age minimization in real-world applications (e.g. \cite{huang2022contextual,wu2022towards}). In other studies (e.g., \cite{chen2022age,fatale2021regret,song2022regret}), authors proposed the concept of AoI regret, where AoI metrics correspond to rewards in the traditional CMAB problem and the goal of minimizing AoI is transformed into minimizing the cumulative regret in the traditional CMAB problem. To sum up, all the existing studies focused on either the CMAB problem with fairness constraints or the CMAB with regularity guarantees. None of them considered the CMAB problem simultaneously maximizing cumulative rewards while guaranteeing fairness among arms and the short-term reward regularity of each arm.

In this paper, we propose a parameterized regular and fair learning algorithm to achieve the aforementioned three objectives. Specifically, we introduce virtual queues and Time-Since-Last-Reward (TSLR) metrics. Virtual queues are leveraged to track cumulative fairness violations. TSLR is similar to AoI, capturing the elapsed number of rounds since the last round an arm received a reward. We utilize TSLR metrics to capture the reward regularity performance. The proposed algorithm linearly combines virtual queue-lengths, TSLR metrics, and Upper Confidence Bound (UCB) estimates in its weight measure. By leveraging UCB estimates, the algorithm is able to balance the tradeoff between exploration and exploitation in online learning, leading to more effective decision-making. 
Note that \cite{li2019combinatorial} only focused on the reward and fairness metrics, while \cite{li2021efficient} considered the reward and TSLR metrics. Both these works are special cases of our proposed algorithm. More importantly, they cannot infer the performance tradeoff between fairness and TSLR metrics, which was captured in this paper. It is challenging to analyze the performance of our algorithm because of the strong coupling between virtual queue-lengths and TSLR metrics, as well as the abrupt dynamics of TSLR, different from that of the virtual queue-lengths. To address these challenges, we reveal a key relationship between these two metrics and employ several non-trivial Lyapunov functions to conduct their drift analyses. Our contributions are summarized as follows:

$\bullet$ We develop a parameterized regular and fair algorithm that linearly combines virtual queue-lengths, TSLR metrics, and UCB estimates (cf. Section \ref{sec:algorithm}). 

$\bullet$ We reveal the key relationship between virtual queue-lengths and TSLR metrics (cf. Lemma \ref{eqn:lemma:QT:relation}). Then, we utilize an elaborate Lyapunov function to obtain the expected negative drift and the bounded absolute drift. Finally, we show that our proposed algorithm achieves zero cumulative fairness violations after a certain number of rounds, which is characterized in terms of algorithmic parameters (cf. Proposition \ref{prop:zero_violation}).


$\bullet$ We derive an upper bound on the running average of mean TSLR metrics (i.e., short-term reward regularity) (cf. Proposition \ref{prop:age}). The derived upper bound has two parts: 1) the first part directly follows from the upper bound on the total mean virtual queue-lengths from the proof of Proposition \ref{prop:zero_violation} and Lemma \ref{eqn:lemma:QT:relation}; 2) the second part is derived by considering a slightly different Lyapunov function and carefully performing drift analysis.



$\bullet$ We obtain an upper bound on the cumulative regret over consecutive $T$ rounds under our proposed algorithm (cf. Proposition \ref{prop:regret}) by combining the drift-plus-penalty technique and regret analysis method for the classical UCB algorithm.



$\bullet$ We validate our theoretical findings in 
the applications for multi-user interactive and panoramic scene delivery and timely information
delivery via wireless (cf. Section \ref{sim:real_world}).

This work extends INFOCOM 2024 \cite{wu2024infocom} in the following aspects: (1) we conduct a deep literature survey related to our research; (2) more detailed proofs for Proposition \ref{prop:zero_violation}, Proposition \ref{prop:age}, and Proposition \ref{prop:regret} are included; (3) we add additional simulations to demonstrate the superior performance of our algorithm based on the real-world dataset.


\emph{Note on Notation}: We use bold and script font of a variable to denote a vector and a set, respectively. We use $\mb{x}/\mb{y}$ to denote the component-wise division of the vector $\mb{x}$ and $\mb{y}$. We use $\sqrt{\mb{x}}$ to denote the component-wise square root of the vector $\mb{x}$. Let $\|\mb{x}\|_1$ and $\|\mb{x}\|$ denote the $l_1$ and $l_2$ norm of the vector $\mb{x}$, respectively. We use $f(x)=o(g(x))$ to denote $\lim _{x \rightarrow \infty} f(x)/g(x)=0$ and $f(x)=O(g(x))$ to denote $\limsup _{x \rightarrow \infty} f(x)/g(x)<\infty$ for positive functions $f$ and $g$.

\section{related work}
In this section, we provide an overview of two key concepts closely related to our research: multi-armed bandit with fairness constraints and age of information.
\subsection{Multi-Armed Bandit with Fairness Constraints}
The Multi-armed Bandit (MAB) problem is a framework where an agent learns system dynamics while optimizing decisions based on prior learning. 
This framework has been widely applied in various fields for decision making, such as clinical tests, online advertising, and adaptive routing. 
Thus, MAB has received significant interest from researchers (e.g., \cite{lai1985asymptotically,auer2002finite,garivier2011kl,agrawal2012analysis}).
Notably, Lai and Robbins established a key logarithmic lower bound on cumulative regret over finite rounds under a class of uniformly good policies.  
This logarithmic regret bound has been further proved to be achieved by UCB \cite{auer2002finite}, Kullback-Leibler UCB (KL-UCB) \cite{garivier2011kl}, and Thompson sampling \cite{agrawal2012analysis}. Subsequent works proposed a new framework called CMAB (e.g., \cite{gai2012combinatorial,chen2013combinatorial,combes2015combinatorial}), which extends the basic MAB framework by allowing the selection of a subset of arms at each round, rather than just one. CMAB can be adapted to more complex real-world scenarios. For example, \cite{gai2010learning} leveraged CMAB to search for an allocation of channels for all users that maximizes the expected sum throughput, where each arm corresponds to a matching of the users to channels. Then, more works applied the framework of CMAB to more real-world network applications such as resource allocation (e.g., \cite{ortiz2019cbmos}), network routing (e.g., \cite{fu2022optimal}), and wireless user scheduling (e.g., \cite{alipour2022multiuser,kang2020low}). Recently, some studies focused on fairness guarantee, in addition to the objective of maximizing the sum of expected rewards for the CMAB problem. The authors in \cite{li2019combinatorial} introduced virtual queues to track the fairness violation and incorporated them into the algorithm design together with the UCB weight \cite{auer2002finite} for estimating the rewards. They characterized cumulative regret and long-term fairness performance. The authors in \cite{liu2021efficient} proposed a pessimistic-optimistic algorithm that achieves state-of-the-art regret performance and a zero fairness constraint violation by properly selecting algorithmic parameters. The authors in \cite{patil2021achieving} provided a fairness guarantee that holds uniformly over time, allowing any MAB algorithm to be chosen. 
However, all these works didn't address the reward regularity performance. After introducing the regularity metrics, solving the problem becomes even more difficult due to the existence of a strong coupling between fairness constraints and reward regularity.
\subsection{Age of Information}
The concept of reward regularity shares similarities with service regularity (e.g., \cite{li2014throughput,li2015wireless}) or the AoI (e.g., \cite{lu2018age,kadota2018optimizing,kadota2019minimizing} and see \cite{sun2019age} and \cite[Ch. 8] {pappas2022age} for an overview). These studies primarily focused on characterizing either the steady-state service regularity or the average AoI. 
However, they didn't delve into algorithm design in unknown environments, where algorithms must not only make decisions but also learn unknown system statistics. 
Subsequently, studies (e.g., \cite{bhandari2020age,chen2022age,fatale2021regret,song2022regret}) proposed the concept of AoI regret, which is a variant of rewards in traditional CMAB problems. The goal is to minimize the cumulative AoI over finite consecutive rounds.
In some other studies (e.g., \cite{moltafet2020average, mlika2020association}), authors investigated and quantified the fairness that can be achieved while minimizing AoI. However, none of them introduced a virtual queue to track the fairness violations and fully reveal the relationship between fairness and AoI. Furthermore, they also didn't consider the unknown network environments.
In conclusion, existing research has primarily concentrated on addressing the CMAB problem with either fairness constraints or regularity guarantees. However, there is a lack of studies that approach the CMAB problem from the angle of simultaneously maximizing cumulative rewards, ensuring fairness across different arms, and maintaining short-term reward regularity for each arm.

\section{system model}
\label{sec:model}





We consider a combinatorial multi-armed bandit with $N$ arms, where multiple arms can be pulled simultaneously in each round. If arm $n$ is pulled in the $t^{th}$ round, it will receive a reward $X_n(t)$. We assume that $\{X_n(t)\}_{t\geq0}$ are independently and identically distributed (i.i.d.) Bernoulli random variables with unknown mean $\mu_n\in(0,1]$\footnote{Our algorithm and analysis can be extended to other probability distributions with a finite support (e.g., \cite{chen2013combinatorial,chen2016combinatorial}).}. We let $\mu_{\min}\triangleq\min_{n}\mu_n>0$ and $\mu_{\max}\triangleq\max_{n}\mu_n\leq1$. Let $S_n(t)=1$ if arm $n$ is pulled in round $t$, and $S_n(t)=0$ otherwise. Hence, the received reward $R(t)$ in round $t$ can be expressed as $R(t)\triangleq\sum_{n=1}^{N}X_n(t)S_n(t)$. Let $\mb{S}(t)\triangleq(S_n(t))_{n=1}^{N}$ be the arm \emph{activation vector}. With a little bit of abuse of notation, we also treat $\mb{S}(t)$ as a set of arms that can be pulled simultaneously in round $t$. We use $\mc{S}$ to denote the collection of all arm activation vectors. Let $S_{\max}$ be the maximum number of arms that can be pulled simultaneously in each round. 

We aim to achieve the following three goals simultaneously: 1) maximizing the \emph{expected cumulative reward} {over consecutive $T$ rounds} (i.e., $\sum_{t=0}^{T-1}\bE[R(t)]$); 2) ensuring \emph{fairness} among arms (i.e.,  a minimum amount of expected reward received by each arm on average); 3) guaranteeing the \emph{reward regularity} of each arm (i.e., how often each arm receives the reward). 
Here, the fairness means that each arm $n$ is at least received the reward $\lambda_n>0$ on average, i.e., 
\begin{align*}
\liminf_{T\rightarrow\infty}\frac{1}{T}\sum_{t=0}^{T-1}\bE[X_n(t)S_n(t)]\geq\lambda_n, \forall n=1,2,\ldots,N,    
\end{align*}
where we assume that $(\lambda_1,...,\lambda_n)$ is feasible (see \cite{li2019combinatorial,liu2021efficient}) in the sense that the system can provide fairness guarantees under some algorithm.
If the statistics of rewards (i.e., $\bs{\mu}\triangleq(\mu_n)_{n=1}^{N}$) are known in advance, then the first two goals can be achieved by deploying a randomized stationary strategy $\{q^{*}(\mb{S}),\forall\mb{S}\in\mc{S}\}$, where $q^{*}(\mb{S})$ is the probability of pulling a set $\mb{S}$ of arms and solves the following optimization problem:
\begin{align}
\max_{q(\mb{S})} &\quad \sum_{\mb{S}\in\mc{S}}q(\mb{S})\sum_{n=1}^{N}\mu_nS_n \\
s.t. &\quad \lambda_n+\delta\leq\sum_{\mb{S}\in\mc{S}}q(\mb{S})S_n\mu_n, \forall n=1,2,\ldots,N, \label{constraint(2)}
\end{align}
where $\delta>0$ is a ``tightness" constant, and $\lambda_n+\delta\leq\mu_n\leq1$ due to the fact that $S_n\leq1$. However, the statistics of rewards are unknown in practice. Hence, the algorithm needs to quickly learn these statistics 
(also known as (a.k.a.) exploration) while pulling arms with the largest empirical rewards so far (a.k.a. exploitation). Note that maximizing the expected cumulative rewards is equivalent to minimizing the cumulative regret over consecutive $T$ rounds, defined as the gap between the expected accumulated reward and the optimal expected reward, i.e.,
\begin{align*}
\text{Reg}(T) \triangleq \sum_{t=0}^{T-1}\sum_{n=1}^{N}\mu_n\left(\bE[S^*_n]-\bE\left[S_n(t)\right]\right),
\end{align*}
where $\bE[S^*_n]\triangleq\sum_{\mb{S}\in\mc{S}}q^*(\mb{S})S_n, \forall n$.

To address our third goal, i.e., quantifying the reward regularity of each arm, we introduce a counter $Z_n(t)$, namely \emph{Time Since Last Reward} (TSLR)\footnote{Here, the TSLR metric is essentially the same as the time since the last service (e.g., \cite{li2015wireless,li2014throughput}) and age of information (e.g., \cite{lu2018age,kadota2019minimizing,sun2019age}).}, to denote the elapsed number of rounds since the last round arm $n$ received the reward until round $t$. Specifically, $Z_n(t)$ increases by one if arm $n$ does not receive the reward in round $t$, either because it is not pulled (i.e., $S_n(t)=0$) or because its reward is zero (i.e., $X_n(t)=0$), and resets to one otherwise, i.e., 
\begin{align}
\label{eqn:age:dynamics}
Z_n(t+1) = 
\begin{cases}
Z_n(t) + 1  & \text{if $S_n(t)X_n(t)=0$}; \\
1           & \text{if $S_n(t)X_n(t)=1$}.
\end{cases}
\end{align}
Hence, the TSLR $Z_n(t)$ captures the ``reward age'' of arm $n$ since the last round receiving the reward and is closely related to the inter-reward interval. Indeed, by following the exact same argument in \cite{li2015wireless}, we can show that the normalized second moment of the inter-reward interval of each arm is proportional to the mean value of its TSLR. Thus, the smaller the TSLR, the more regularly the arm receives the reward. As such, the third goal is equivalent to minimizing the running average of total expected TSLR metrics over consecutive $T$ rounds, i.e., $\frac{1}{T}\sum_{t=0}^{T-1}\sum_{n=1}^{N}\bE[Z_n(t)]$.

\section{Algorithm Design and Analysis}
\label{sec:algorithm}
In this section, we achieve the aforementioned triple objectives by developing a parametric class of algorithms that efficiently utilize a combination of UCB estimates for minimizing the cumulative regret (see \cite{auer2002finite}), TSLR metrics measuring the reward regularity (see \cite{li2014throughput, li2015wireless, li2017emulating}), and virtual queues addressing the fairness among arms (see \cite{neely2010stochastic} for an overview) in their decisions. Here, the UCB weights are utilized to balance the exploitation-exploration tradeoff in online learning with the goal of achieving minimum cumulative regret. TSLR metrics are introduced to capture each arm's ``reward age'' to guarantee that it receives a reward regularly. Virtual queues are used to track the ``reward debt'' and thus the cumulative fairness violations.

In order to obtain the UCB weight, we define the following notations. Let $H_n(t)$ be the number of rounds arm $n$ has been pulled until round $t$, i.e., $H_n(t)\triangleq\sum_{\tau=0}^{t-1}S_n(\tau)$. We set $H_n(0)=0$ due to the fact that the system starts at $t=0$. We use $\ol{\mu}_n(t)$ to denote the sample mean of the received rewards of arm $n$ until round $t$, i.e., $\ol{\mu}_n(t)\triangleq\left(\sum_{\tau=0}^{t-1}X_n(\tau)S_n(\tau)\right)/H_n(t)$. If $H_n(t)=0$ (i.e., arm $n$ has not been pulled yet until round $t$), we set $\ol{\mu}_n(t)=1$. Let $w_n(t)$ denote the UCB weight of arm $n$ in round $t$ and be defined as follows:
\begin{align}
w_n(t)\triangleq \min\left\{\ol{\mu}_n(t)+\sqrt{\frac{3\log t}{2H_n(t)}}, 1\right\},
\end{align}
where $\sqrt{3\log t/(2H_n(t))}$ is the exploration term that quantifies the uncertainty of the sample mean $\ol{\mu}_n(t)$. A smaller $H_n(t)$ implies less exploration on arm $n$ and thus less accuracy of its sample mean estimation. In such a case, arm $n$ should get a higher priority to be pulled. Here, we use the truncated version of the UCB weight, since the actual reward of each arm is at most $1$. Again, when $H_n(t)=0$, we set $w_n(t)=1$, i.e., if arm $n$ has not been pulled yet until round $t$, it has the highest priority to pull. 

To address fairness among arms, we introduce a virtual queue for each arm to keep track of its ``reward debt'' over rounds. In particular, we use $Q_n(t)$ to denote the virtual queue-length of arm $n$ at the beginning of round $t$, which evolves as follows:
\begin{align}
\label{eqn:virtualQ}
Q_n(t+1)=(Q_n(t)+\lambda_n-S_n(t)X_n(t)+\epsilon)^{+},
\end{align}
where $(x)^{+}\triangleq\max\{x,0\}$ and $\epsilon\in(0,1)$ is some positive parameter that ensures $\lambda_n+\epsilon<\mu_n\leq1, \forall n$.
We set $Q_n(0)=0,\forall n$ as the system starts at $t=0$. 

In order to achieve a low cumulative regret, we would like to pull arms with large UCB estimates in each round. In particular, we want to pull arms with high sample mean rewards as well as arms with large uncertainties of received rewards due to fewer explorations. To ensure the reward regularity of each arm, we also need to pull arms with large TSLRs. Moreover, to guarantee desired fairness among arms, arms with large virtual queue-lengths should get high priorities to be pulled. This naturally motivates the following algorithm. 


\begin{algorithm}
\caption{Regular and Fair Learning (RFL) Algorithm}
In round $t$, pull a set of arms  $\wh{\mb{S}}(t)\triangleq(\wh{S}_n(t))_{n=1}^{N}$ satisfying
\begin{align*}
\wh{\mb{S}}(t)\in\argmax_{\mb{S}\in\mc{S}}\sum_{n=1}^{N}\left(Q_n(t)+\alpha Z_n(t)+\beta w_n(t)\right)S_n,
\end{align*}
\begin{flushleft}
where $\alpha \geq 0 $ and $\beta \geq 0$ are control parameters. Then, update the TSLR metrics $\mb{Z}(t)\triangleq(Z_n(t))_{n=1}^{N}$ according to \eqref{eqn:age:dynamics} and virtual queue-lengths $\mb{Q}(t)\triangleq(Q_n(t))_{n=1}^{N}$ according to \eqref{eqn:virtualQ}.
\end{flushleft}
\end{algorithm}

In our proposed RFL algorithm, parameters $\alpha$ and $\beta$ can be adjusted to balance the TSLR metrics and the UCB estimates. When $\alpha=0$, our RFL algorithm coincides with fair learning algorithms that aim to achieve near-optimal cumulative regret while guaranteeing fairness among arms (e.g., \cite{li2019combinatorial,liu2021efficient}). As $\alpha$ increases, our algorithm puts more weight on TSLR metrics and thus results in better reward regularity performance. When $\beta=0$, our RFL algorithm reduces to the algorithms that aim to balance the (virtual) queue-lengths and regularity performance in steady-state in the context of wireless scheduling (e.g., \cite{li2014throughput, li2015wireless, li2017emulating}). A larger $\beta$ emphasizes more on the UCB estimates and thus yields a smaller cumulative regret. 

Next, we will analyze fairness, 
reward regularity, and cumulative regret performance under our proposed RFL algorithm. The main challenge lies in the strong coupling between the virtual queue-lengths and TSLR metrics and the abrupt dynamics of TSLR metrics. Indeed, if an arm does not receive the reward in one round, then both its virtual queue length and TSLR increase. Otherwise, the virtual queue length decreases by a finite amount, and the TSLR resets to one. Moreover, the TSLR metric increases at most by one and has an unbounded decrement if the corresponding arm receives the reward, which is significantly different from the evolution of virtual queue-lengths. 
As such, we first reveal the key relationship between the virtual queue-length and TSLR metric of each arm, as shown in the following lemma.





\begin{lemma}
\label{eqn:lemma:QT:relation}
For each arm $n$, if $Q_n(0)=Z_n(0)=0$, then 
\begin{align}
1+Q_n(t)\geq \lambda_n Z_n(t), \quad\forall t\geq0,
\end{align}
holding for any sample path. 
\end{lemma}
\begin{IEEEproof}
First, we note that $1+Q_n(0)\geq\lambda_n Z_n(0)$ by the initial condition. Suppose that $1+Q_n(t)\geq\lambda_n Z_n(t)$ is true for some $t\geq0$. Then, we have the following two cases:
\begin{itemize}
\item [(i)] If arm $n$ receives a reward in round $t$ (i.e., $\wh{S}_n(t)X_n(t)=1$), then we have $Z_n(t+1)=1$ and thus $1+Q_n(t+1)\geq \lambda_n Z_n(t+1)$ trivially holds since the virtual queue-length is non-negative by its definition and $\lambda_n\in(0,1]$.
\item [(ii)] If arm $n$ doesn't receive a reward in round $t$ (i.e., $\wh{S}_n(t)X_n(t)=0$), then $Z_n(t+1)=Z_n(t)+1$ by the definition of the age and $Q_n(t+1)=Q_n(t)+\lambda_n+\epsilon$. Hence, we have 
\begin{align}
1+Q_n(t+1)=&1+Q_n(t)+\lambda_n+\epsilon\nonumber\\
\geq&\lambda_n Z_n(t)+\lambda_n=\lambda_n Z_n(t+1),
\end{align}
where the second last step follows from the assumption that $1+Q_n(t)\geq\lambda_n Z_n(t)$.
\end{itemize}
Hence, we have $1+Q_n(t+1)\geq\lambda_n Z_n(t+1)$ holding in both cases and hence by using the mathematical induction, we have the desired result.
\end{IEEEproof}

Based on Lemma \ref{eqn:lemma:QT:relation}, at first glance, the fair learning algorithm studied in \cite{liu2021efficient,li2019combinatorial} (cf. RFL algorithm with $\alpha=0$) can provide an upper bound on the expected virtual queue-length in any round $t$ and thus can guarantee the reward regularity performance. This has also been observed in \cite[Proposition 2]{kadota2019scheduling} that captures the long-term fairness and reward regularity performance in our context. However, the weighting parameter $\alpha$ plays a significant role in the short-term performance such as the cumulative fairness violation, reward regularity, and cumulative regret performance, as revealed in our analyses and simulations. 

Noting that the TSLR metric can change abruptly, and is quite different from the virtual queue-length evolution. This requires the careful selection of the Lyapunov function to ensure the proper fairness violation bounds, which heavily rely on \cite[Lemma 2.2]{hajek1982hitting} requiring that the absolute drift of the Lyapunov function is bounded or exponentially decays. This can be corroborated by a counterexample in \cite[Fig. 5]{li2015wireless} that constructs a Markov chain, where there exists a Lyapunov function with a strictly negative expected drift. However, its Lyapunov drift has bounded increments but
unbounded decrements, similar to our TSLR metric, and its mean state value does not exist, let alone its moment-generating function. Despite using a similar Lyapunov function as in \cite{li2015wireless}, we aim to characterize short-term performance instead of long-term or steady-state analysis as in \cite{li2014throughput,li2015wireless}. Under our proposed RFL algorithm, by selecting appropriate $\epsilon$ in the virtual queue evolution (cf. \eqref{eqn:virtualQ}), zero cumulative fairness violation can be achieved after a certain number of rounds, namely \emph{zero-violation point}, as shown in the following proposition. To accommodate the limited space, we provide only sketches of proofs for all our propositions.

\begin{proposition} 
\label{prop:zero_violation}
[Zero Cumulative Fairness Violations] Under the RFL algorithm, if $\epsilon \leq \delta/2$, there exists a zero violation point $t_0\triangleq g_0(\alpha,\beta)/\epsilon=O((\alpha^2\log\alpha+\beta)/\epsilon)$ after which the zero cumulative fairness violation is achieved for any round $t\geq t_0$, i.e., 
\begin{align*}
\sum_{n=1}^{N}\left(\sum_{\tau=0}^{t-1}(\lambda_n-\bE[\wh{S}_n(\tau)X_n(\tau)])\right)^{+}=0, \quad\forall t\geq t_0,
\end{align*}
where $g_0(\alpha,\beta)\triangleq\sqrt{N}\bigg(\frac{1}{\theta(\alpha)}\log(v_0(\alpha)+1)+D(\alpha)+U(\alpha,\beta) \bigg)=O(\alpha^2\log\alpha+\beta)$, $D(\alpha)\triangleq\left(12\alpha+1\right)N/(\lambda_{\min}\mu_{\min})$, $U(\alpha,\beta)\triangleq8N^2(4\alpha+3\beta+2)/(\delta\mu_{\min}^2)$, $\theta(\alpha) \triangleq 3\delta \mu_{\min}/(48ND^2(\alpha)+\delta\mu_{\min}D(\alpha))$,  $v_0(\alpha)\triangleq32N/(\delta \mu_{\min}\theta(\alpha))$, and $\lambda_{\min}\triangleq\min_{n}\lambda_n>0$.
\end{proposition}

\emph{Proof:}  We select the Lyapunov function
\begin{align}
V(t) \triangleq \|\mb{W}(t)\|_2,
\end{align}
where $\mb{W}(t)\triangleq (\mb{Q}(t)/\sqrt{\bs{\mu}},2\sqrt{\alpha\mb{Z}(t)/\bs{\mu}})$. We first establish that the Lyapunov function has an expected negative drift when $V(t)$ is large enough, and its drift is absolutely bounded. Here, the TSLR metrics have abrupt dynamics and need to be handled through non-trivial bounding techniques. Then, we derive an upper bound for $\bE[V(t)]$ by following \cite[Lemma 11]{liu2021efficient}. This together with the fact that $V(t)\geq\|\mb{Q}(t)\|_2\geq\|\mb{Q}(t)\|_1/\sqrt{N}$ results in an upper bound for $\bE\left[\|\mb{Q}(t)\|_1\right]$. Finally, according to the dynamics of virtual queue-lengths, the cumulative fairness violations can be upper bounded by $\bE\left[\|\mb{Q}(t)\|_1\right]$ and thus implies the desired result. Please see Appendix \ref{App:proof:zero_violation} for detailed proof.

\begin{remark}
From Proposition \ref{prop:zero_violation}, we can see that the zero-violation point is inversely proportional to parameter $\epsilon $ used in the virtual queue-length evolution (cf. \eqref{eqn:virtualQ}). This intuitively makes sense since a large $\epsilon$ results in large virtual queue-lengths, enforcing the RFL algorithm to pull arms with large virtual queue-lengths, and thus the system achieves zero cumulative fairness violation faster. In addition, we can observe from Proposition \ref{prop:zero_violation} that a large parameter $\alpha$ or $\beta$ postpones the zero violation point, which also matches our intuition. Indeed, a large $\alpha$ or $\beta$ puts more weight on TSLR metrics or UCB estimates, under which the RFL algorithm pulls arms with larger TSLR metrics or UCB estimates and achieves the zero cumulative fairness violation slower. Moreover, we can see that parameter $\alpha$ has a larger impact on the zero violation point than parameter $\beta$. This is because the increase of the TSLR metric is at least one while the UCB estimate is at most one, and thus the zero violation point is more sensitive to the change of parameter $\alpha$. 
\end{remark}


Next, we characterize the short-term reward regularity performance, which is quite different from the steady-state regularity performance studied in \cite{li2014throughput,li2015wireless}. Note that the proof of Proposition \ref{prop:zero_violation} results in an upper bound for $\bE[\|\mb{Q}\|_1]$, which, together with Lemma \ref{eqn:lemma:QT:relation}, provides an upper bound for the expected TSLR metrics in any round $t$. However, this upper bound increases with respect to parameter $\alpha$, which becomes quite loose especially when $\alpha$ is large. Indeed, as we mentioned before, a larger $\alpha$ puts a larger weight on the TSLR and thus yields a better reward regularity performance. Hence, such a derived upper bound is too loose to quantify the reward regularity performance of our proposed RFL algorithm when $\alpha$ is large. As such, we use a slightly different Lyapunov function and derive an upper bound for the running average of expected TSLR, which is inversely proportional to $\alpha$ when $\alpha$ is not too large, matching our intuition for our RFL algorithm. 

\begin{proposition}
\label{prop:age}
[Short-term Reward Regularity] Under the RFL algorithm, if $\epsilon\leq\delta/2$, then the running average of total expected TSLR metrics over consecutive $T$ rounds can be bounded as follows:
\begin{align*}
\frac{1}{T}\sum_{t=0}^{T-1}\sum_{n=1}^{N}\bE[Z_n(t)] 
\leq \min\bigg\{&\frac{N+g_0(\alpha,\beta)}{\lambda_{\min}}, \\
& \frac{N^2}{\delta\mu_{\min}} \left( 1 + \frac{3\beta+4}{\alpha} \right)\bigg\}.
\end{align*}
Here, $g_0(\alpha,\beta)=O(\alpha^2\log\alpha+\beta)$ is defined in Proposition \ref{prop:zero_violation}.
\end{proposition}

\emph{Proof:} 
The first part of the upper bound directly follows from Lemma \ref{eqn:lemma:QT:relation}  and the upper bound for $\bE[\|\mb{Q}(t)\|_1]$ derived in the proof of Proposition \ref{prop:zero_violation}. However, as we mentioned before, such upper bound increases with respect to parameter $\alpha$, and thus becomes quite loose when $\alpha$ is large. Hence, we need a tighter bound when $\alpha$ is large. To that end, different from the proof of Proposition \ref{prop:zero_violation}, 
we consider the following Lyapunov function
\begin{align}
V_1(t) \triangleq V^2(t)=\|\mb{W}(t)\|_2^2,
\end{align}
where we recall that $\mb{W}(t)\triangleq (\mb{Q}(t)/\sqrt{\bs{\mu}},2\sqrt{\alpha\mb{Z}(t)/\bs{\mu}})$.
After obtaining its negative expected drift, we derive an upper bound on the running average of mean TSLR metrics using telescoping techniques as in the classical Lyapunov drift analysis. Please see Appendix \ref{APP:proof:age} for detailed proof.
\begin{remark}
From the second part of the derived upper bound on the reward regularity in proposition \ref{prop:age}, we can see that the reward regularity performance gets worse as parameter $\beta$ increases. This matches our intuition that the larger parameter $\beta$, the larger the UCB weight compared to the TSLR metrics, degrading the reward regularity performance. In contrast, our derived reward regularity performance is inversely proportional to parameter $\alpha$. This is because a large parameter $\alpha$ puts more emphasis on the TSLR metrics, improving the reward regularity performance, as observed before. Moreover, as $\alpha$ increases to infinity (i.e., $\alpha\uparrow\infty$), the reward regularity is bounded by a constant. This makes sense since the RFL algorithm with an extremely large $\alpha$ serves arms with the largest TSLR metrics. When at most one arm is pulled in each round, it has a similar behavior with the Round-robin algorithm under which the total expected TSLR metrics is constant since the TSLR vector in each round should be $(1, 2, \ldots, N-1)$ and the total sum is equal to $N(N-1)/2$. However, when $\alpha$ decreases to 0 (i.e., $\alpha\downarrow 0$), the second part increases to infinity, and thus the short-term reward regularity is bounded by the first part, which is dominated by parameter $\beta$ with $\alpha\downarrow 0$.
\end{remark}

Lastly, we analyze the cumulative regret performance of our RFL algorithm. 


\begin{proposition}
\label{prop:regret}
[Cumulative Regret] Under the RFL algorithm with $\epsilon\leq\delta$, the cumulative regret $\text{Reg}(T)$ over consecutive $T$ rounds can be bounded from above as follows:
\begin{align*}
\text{Reg}(T)\leq \min \bigg\{&S_{\max}\mu_{\max}T, \frac{NT}{\mu_{\min}} \left(\frac{\alpha + 1}{\beta }\right) \nonumber\\
&+2\sqrt{6NS_{\max}T\log T} +N\left(1+\frac{5\pi^2}{12}\right) \bigg\}.
\end{align*}
\end{proposition}

\emph{Proof:}
The cumulative regret is obviously bounded by linear regret $S_{\max}\mu_{\max}T$, since at most $S_{\max}$ arms can be pulled in each round. Next, we mainly focus on the derivation of logarithmic regret upper bound. To that end, we select the Lyapunov function $L(t)\triangleq\frac{1}{2}\sum_{n=1}^{N}Q_n^2(t)/\mu_n +\alpha \sum_{n=1}^{N} Z_n(t)/\mu_n$
and perform drift-plus-penalty analysis on 
\begin{align}
\bE\left[L(t+1)-L(t)\right]+\beta\Delta R(t),
\end{align}
where $\Delta R(t)\triangleq\sum_{n=1}^{N}\bE\left[\mu_nS_n^*(t)-\mu_n\wh{S}_n(t)\right]$ and the cumulative regret $\text{Reg}(T)\triangleq\sum_{t=0}^{T-1}\Delta R(t)$. Then, we carefully incorporate the regret analysis for the classical UCB algorithm (e.g., \cite{auer2002finite}) into our analysis. The analysis is similar to the line of regret analysis in \cite{hsu2018integrate,li2019combinatorial,liu2021efficient,li2021efficient}, and is available in Appendix \ref{App:proof:regret} for the detailed proof. 

\begin{remark}
The second part of the derived upper bound on the cumulative regret consists of two terms: (i) $2\sqrt{6NS_{\max}T\log T}+N(1+5\pi^2/12)$ has the same order $O(\sqrt{NT\log T})$ as the instance-independent upper bound for the classical UCB algorithm (see \cite[Ch. 2.4.3]{bubeck2012regret}) without any fairness constraints and thus this term is attributed to the cost involved in the exploration/exploitation process in online learning; (ii) $NT(\alpha + 1)/(\mu_{\min}\beta) $ decreases as parameter $\alpha$ decreases and parameter $\beta$ increases. This also matches our intuition on the RFL algorithm: a smaller $\alpha$ or a larger $\beta$ means that our algorithm puts less weight on the TSLR metric or more weight on the UCB estimates, which makes arms with larger UCB weights pulled more often, yielding a smaller cumulative regret. However, as $\alpha$ increases to infinity or $\beta$ decreases to $0$, the second part also increases to infinity and thus the cumulative regret is bounded by a constant linear bound in the first part of our derived upper bound. 
\end{remark}

\begin{figure}[htbp]
\centering 
\vspace{-0.2in}
\includegraphics[scale=0.33]{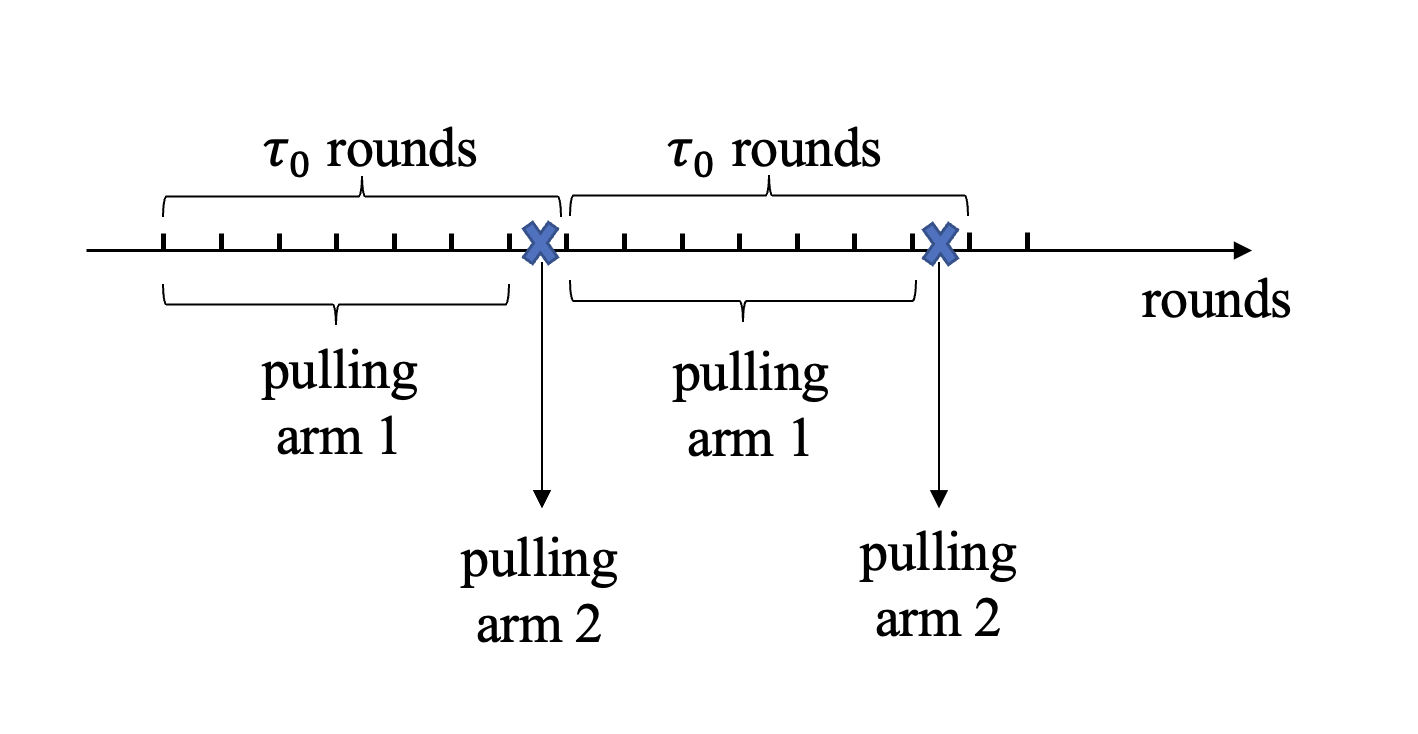}
\caption{Arm pulling schedule.}
\label{fig:remark_tau0}
\end{figure}

\begin{remark} \label{remark:tradeoff}
When $\alpha$ is not too large, the upper bounds derived in Proposition \ref{prop:age} and Proposition \ref{prop:regret} are dominated by their second part and reveal a fundamental tradeoff: when increasing $\beta/\alpha$, the cumulative regret improves, but the short-term reward regularity performance deteriorates. That is, the improvement of the cumulative regret is at the cost of degrading the reward regularity performance. Such a tradeoff might be tight in some cases, e.g., $\beta>\alpha$. Considering there are two arms, at most one arm can be pulled in each round. Suppose $\mu_1 > \mu_2$, and assume that both arms are pulled sufficiently many times. In this case, their UCB weight $w_1(t)$ and $w_2(t)$ are very close to their true mean $\mu_1$ and $\mu_2$. Under our algorithm, arm 2 is pulled roughly once every $\tau_0=\lceil(\alpha+\beta(\mu_1-\mu_2))/(\lambda_2+\alpha)\rceil=O(\beta/\alpha)$ rounds, and arm 1 is pulled in all other rounds under our proposed RFL algorithm, as shown in Fig. \ref{fig:remark_tau0}. Indeed, if arm 1 is pulled, then the weight of arm 2 increases by $\alpha+\lambda_2$ while the weight of arm 1 roughly remains the same (i.e., $\alpha+\beta\mu_1$ due to its virtual queue-length being $0$ and TSLR metric being $1$) until $(\alpha+\lambda_2)\tau_0+\beta\mu_2>\alpha+\beta\mu_1$. Therefore, the running average TSLR metric of arm 2 is roughly equal to $(1+2+3+. . .+ \tau_0)/\tau_0= (\tau_0+1)/2$. 
Meanwhile, the running average TSLR metric of arm 1 is roughly equal to $ (\tau_0-1+2)/\tau_0 = 1+ 1/\tau_0$. As such, the total running average of TSLR metrics is $O(\beta/\alpha)$. On the other hand, the cumulative regret is roughly equal to $(\mu_1-\mu_2)T/\tau_0 = O(T\alpha/\beta)$.
\end{remark}


Table \ref{table:tradeoff} provides three typical sets of parameters $(\alpha,\beta,\epsilon)$ and their impacts on cumulative fairness violations, reward regularity, and cumulative regret performance, characterized by Proposition \ref{prop:zero_violation}, Proposition \ref{prop:age}, and Proposition \ref{prop:regret}, respectively. In particular, we set $\epsilon=O(1/\sqrt[6]{T})$ to ensure that $\epsilon<\delta/2$ for a large time horizon $T$ and provide three different sets of $\alpha$ and $\beta$ values to illustrate the tradeoff among cumulative regret and short-term reward regularity while achieving zero cumulative fairness violations after a constant number of rounds. First, note that according to the evolution of virtual queue-lengths (cf. Proposition \ref{prop:zero_violation}), the zero cumulative fairness violation can be achieved only when $\bE\left[\|\mb{Q}(t)\|_1\right]$ is on the order of $\epsilon t$. Meanwhile, $\bE\left[\|\mb{Q}(t)\|_1\right]$ is dominated by $O(\alpha^2\log \alpha+\beta)$ according to the upper bound of $\bE\left[\|\mb{Q}(t)\|_1\right]$ (cf. Proof of Proposition \ref{prop:zero_violation}). Hence, if $\alpha^2\log \alpha+\beta=o(\epsilon t)$, i.e., $(\alpha^2\log \alpha+\beta)/\epsilon=o(t)$, then, zero violation point is sublinear with respect to the time horizon $T$ (i.e., $(\alpha^2\log \alpha+\beta)/\epsilon=o(T)$), guaranteeing  
zero cumulative fairness violations. Next, we analyze the performance of reward regularity and cumulative regret with different $\alpha$ and $\beta$ values:

\begin{itemize}
    \item [(1)] When $\beta = 0$, our RFL algorithm only contains virtual queues and TSLR metrics, which is similar to that studied in \cite{li2014throughput,li2015wireless}. However, \cite{li2014throughput,li2015wireless} only focus on throughput-optimality and service regularity in steady-state while we are interested in short-term performance, e.g., cumulative fairness violations and short-term reward regularity. With $\beta=0$, we set $\alpha=(\sqrt[6]{T})$ to ensure that the zero violation point is sublinear, e.g., $t_0=O(\sqrt{T}\log T)$. In such a case, the reward regularity is bounded by some constant while the cumulative regret is linear, i.e., $O(T)$. 
\item[(2)] When $\alpha=0$, our RFL algorithm reduces to the fair learning algorithm in \cite{liu2021efficient,li2019combinatorial}. We set $\beta=O(\sqrt{T})$ and thus the zero violation point $t_0$ becomes $O(\sqrt[3]{T^2})$, which is still sublinear and guarantees zero cumulative fairness violations. In such a case, our algorithm achieves the same order-wise cumulative regret performance as in \cite{liu2021efficient,li2019combinatorial} while guaranteeing the reward regularity $O(\sqrt{T})$. This is because the total expected virtual queue-lengths is $O(\sqrt{T})$ and the reward regularity performance follows by Lemma \ref{eqn:lemma:QT:relation}. Interestingly, the algorithm involving only TSLR metrics and UCB estimates (see \cite{li2021efficient}) achieves the same cumulative regret and short-term reward regularity performance as our RFL algorithm, implying that the TSLR metrics behavior similarly to the virtual queue-lengths in terms of algorithm operations together with the UCB estimates. 
\item [(3)] Compared with the second case, we keep $\beta=(\sqrt{T})$ and $\epsilon=O(1/\sqrt[6]{T})$ unchanged and increase $\alpha$ from 0 to $O(\sqrt[6]{T})$. Interestingly, the order of the zero violation point does not change, and thus the zero cumulative fairness violations are still achieved. In addition, the reward regularity performance improves from $O(\sqrt{T})$ to $O(\sqrt[3]{T})$. This is at the cost of deteriorating cumulative regret performance from $O(\sqrt{T\log T})$ to $O(\sqrt[3]{T^2})$. The choice of parameter $\alpha$ provides the flexibility of trading off reward regularity and cumulative regret performance and adapting to different application scenarios. 
\end{itemize}


\begin{table}[h!]\footnotesize
\renewcommand\arraystretch{1.3}
\centering
\scalebox{0.8}{
\begin{tabular}{|l|c|c|c|}
\hline
\diagbox{$(\alpha,\beta,\epsilon)$} {Performance}& Zero Violation Point & Regularity & Regret \\
\hline
$(O(\sqrt[6]{T}), \beta=0,O(1/\sqrt[6]{T}))$ &$O(\sqrt{T}\log T)$ & $O(1) $& $O(T)$ \\
\hline
$(\alpha=0,O(\sqrt{T}),O(1/\sqrt[6]{T}))$ &$O(\sqrt[3]{T^2})$ & $O(\sqrt{T}) $& $O(\sqrt{T\log T})$ \\
\hline
$(O(\sqrt[6]{T}),O(\sqrt{T}),O(1/\sqrt[6]{T}))$ &$O(\sqrt[3]{T^2})$ & $O(\sqrt[3]{T}) $& $O(\sqrt[3]{T^2})$ \\
\hline 
\end{tabular}
}
\caption{Performance tradeoff}
\label{table:tradeoff}
\end{table}
\vspace{-0.2in}

\section{Motivating Applications}
\label{sec:motivating}
In this section, we illustrate two motivating applications for regular and fair learning in the CMAB framework: (i) interactive and panoramic scene delivery over wireless networks and (ii) timely information delivery over wireless networks. 

\begin{figure}[htbp]
\centering 
\vspace{-0.1in}
\subfloat[Panoramic Scene Delivery]{
\label{fig:app_panoramic}
\hspace{-0.15in}\includegraphics[width=1.8in,height=1.2in]{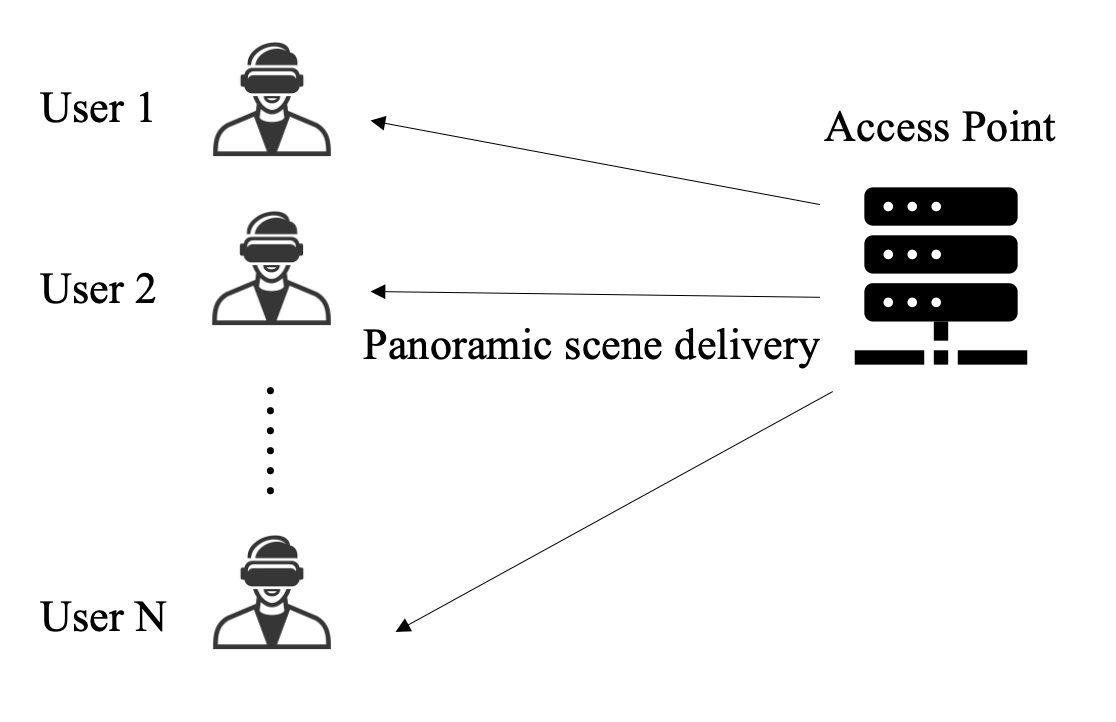}
\hspace{-0.25in}
} 
\subfloat[Timely Information Delivery]
{ \label{fig:app_AoI} 
\includegraphics[width=1.8in,height=1.2in]{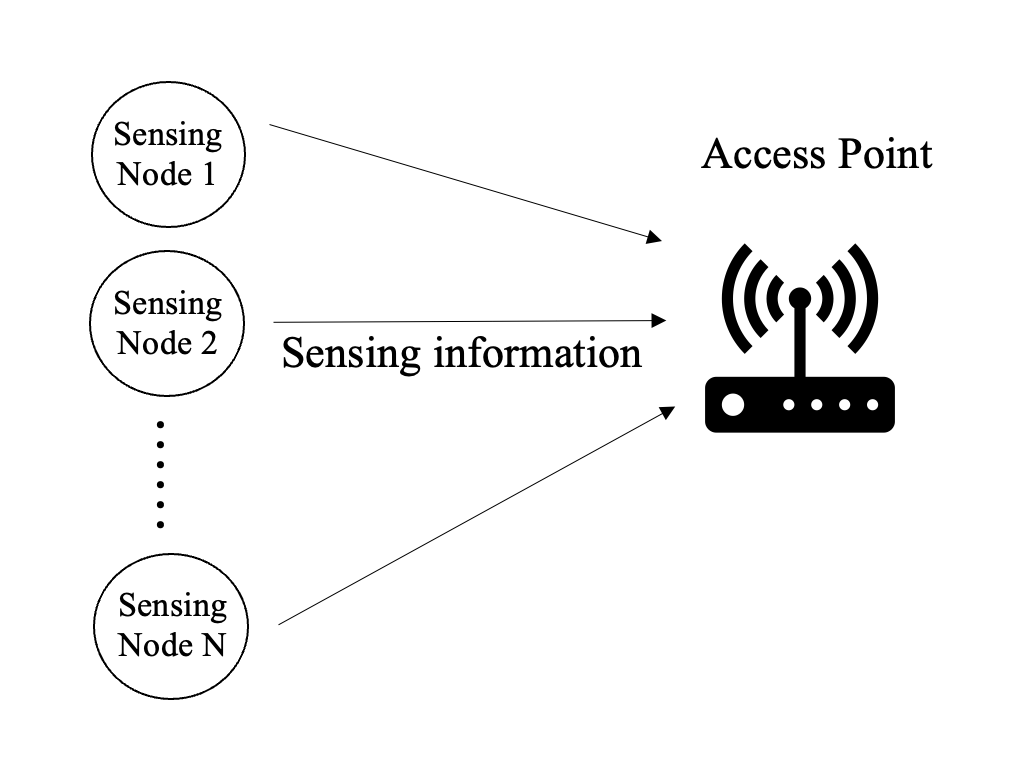}
\hspace{-0.2in}
} 
\caption{Motivating applications.}
\label{fig:app}
\vspace{-0.3in}
\end{figure}

\subsection{Interactive and Panoramic Scene Delivery}
\label{app:panoramic}

We consider the problem of delivering interactive and panoramic scenes (e.g., virtual reality) from an access point (AP) to multiple users. We assume that there is no playback buffer on the user’s device to ensure timely and smooth interactions. Note that panoramic scene delivery typically requires $4\sim6\times$ bandwidth than the typical video transmission with the same resolution. Fortunately, a user can only see roughly 20\% of the panoramic content, called \emph{Field of View (FoV)}, thus it is sufficient to deliver FoV if the user's motion can be predicted accurately. However, the motion prediction always incurs an error and thus we typically deliver a portion larger than the FoV to tolerate the predictor error. Given a predicted viewport, the panoramic scene can be partitioned into a finite number of delivery portions. Each delivery portion corresponds to an unknown successful viewing probability, which is the product of the viewport prediction probability and the successful transmission probability. Here, the viewport probability refers to the probability that the delivery portion covers the actual user's FoV, while the successful transmission probability means the probability that the selected portion can be successfully delivered. The larger the delivery portion, the higher the viewport prediction probability and the lower the successful transmission probability. Please see \cite{chen2021motion,chen2020thompson} for more detailed modeling of interactive and panoramic scene delivery for a single user. 

\begin{figure}[htbp]
\centering 
\vspace{-0.2in}
\includegraphics[scale=0.22]{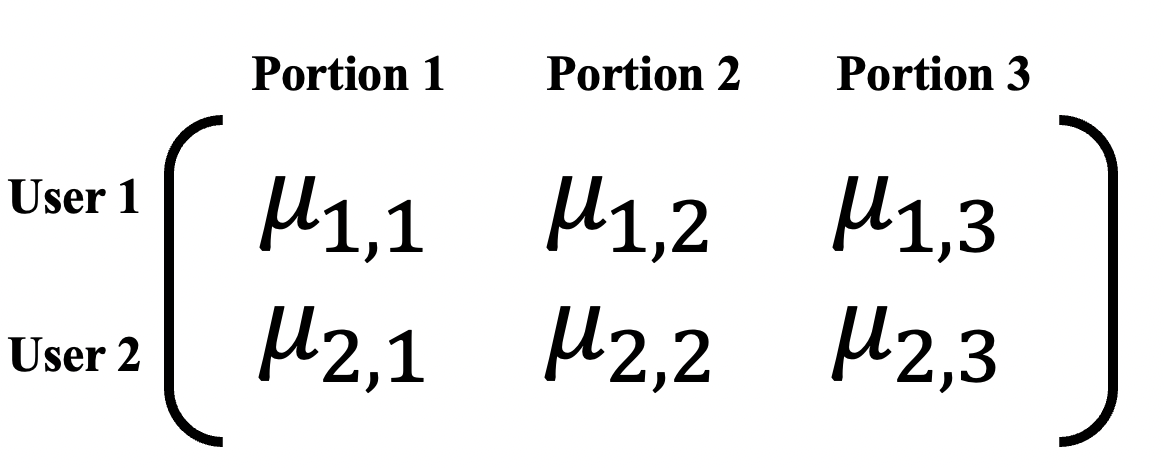}
\caption{User-portion pairs in panoramic scene delivery.}
\label{fig:panoramic_matrix}
\vspace{-0.3in}
\end{figure}

The goal is to maximize the average rate of successfully viewing the content while guaranteeing the minimum required rate for each user. Moreover, we need to provide a seamless user experience (i.e., how often each user gets successful views) subject to wireless interference constraints. The considered problem can be mapped to our regular and fair learning framework, where each arm corresponds to the pair of each user and its selected delivery portion. The difference lies in that fairness refers to guaranteeing the minimum successful content viewing rate for each user and reward regularity also refers to each user how often each user successfully sees the delivered content. Fig. \ref{fig:panoramic_matrix} shows an example with two users, where each user has three different portions to select for wireless transmission. $\mu_{n,m}$ denotes the probability that user $n$ successfully sees the content if delivery portion $m$ is selected and is unknown a priori. Hence, the user scheduler is similar to our RFL algorithm, where the UCB weight of each user is defined as the maximum of the UCB estimates of its all possible delivery portions. Once the user schedule is determined, each selected user selects the delivery portion with the maximum UCB estimate. Our framework can be easily extended to deal with this application, as shown in our simulations (cf. Section \ref{sim:real_world}).

\subsection{Timely Information Delivery via Wireless}
\label{app:rate}
We consider the problem of scheduling multiple sensing sources to transmit sensing information to one AP subject to the wireless interference constraints, where only a subset of sensing sources can transmit data simultaneously. Each node associated with the AP can also select different rates to transmit. The wireless channel is typically unreliable and thus is associated with an unknown successful transmission probability (e.g., \cite{li2021efficient}). To ensure the information freshness at the AP, the age of information (AoI) is typically introduced to measure the time elapsed since the last time the information was successfully delivered. The goal is to maximize the system throughput (i.e., the amount of sensing information successfully delivered to the AP) while guaranteeing fairness among sensing sources (i.e., the minimum amount of successfully delivered information required by each sensing source) and minimizing the average AoI. 

\begin{figure}[htbp]
\centering 
\vspace{-0.7in}
\includegraphics[scale=0.22]{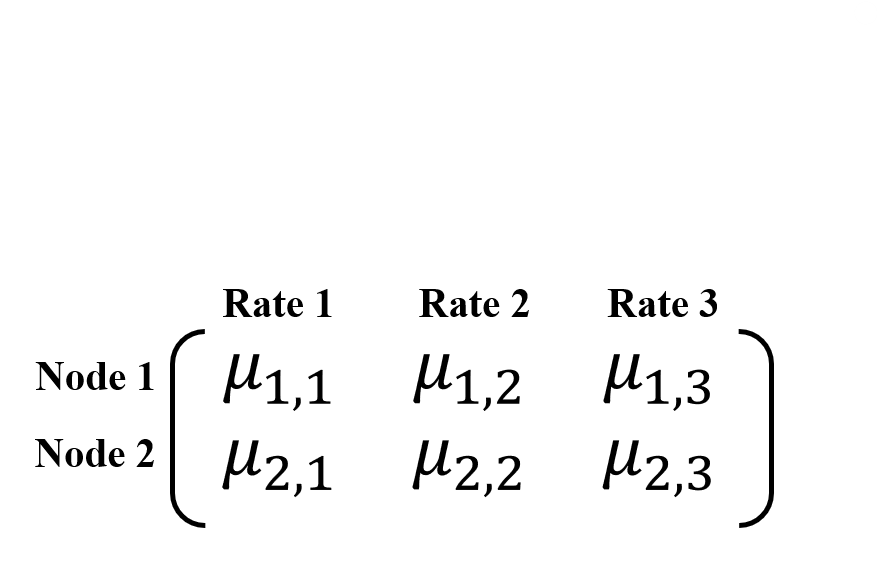}
\vspace{-0.1in}
\caption{Node-rate pairs in timely information delivery.}
\label{fig:rate_matrix}
\vspace{-0.2in}
\end{figure}

This problem can be formulated as our regular and fair learning problem. In particular, each arm corresponds to the pair of each node and its selected transmission rate. AoI is equivalent to the TSLR metric in our formulation. Fig. \ref{fig:rate_matrix} shows an example with two nodes, where each node has three different rates to select for wireless transmission. $\mu_{l,k}$ denotes the probability that node $l$ successfully transmits sensing data to the AP at rate $k$ and is unknown a priori. The scheduling mechanism resembles our RFL algorithm, in which the UCB weight for each node is calculated as the maximum among the UCB estimates for all its feasible transmission rates. Once the node schedule is determined, each selected node selects the transmission rate with the maximum UCB estimate. Our proposed RFL algorithm can be utilized to determine which and when each sensing source should transmit, as well as which rate to select for transmission for the scheduled node to achieve the triple goals, as shown in our simulations (cf. Section \ref{sim:real_world}).

\label{sec:simulation}
\begin{figure*}[!htbp]
\vspace{-0.3in}
\centering 
\subfloat[Average Reward]{
\label{fig:sim:alpha_frac}
\includegraphics[scale=0.24]{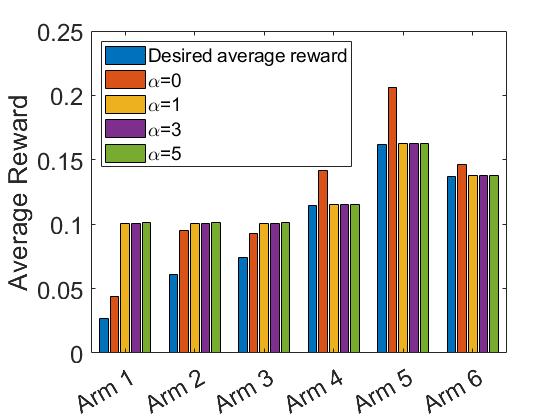}
\hspace{-0.28in}
} 
\subfloat[Cumulative Fairness Violations]{ \label{fig:sim:alpha_violation}
\includegraphics[scale=0.24]{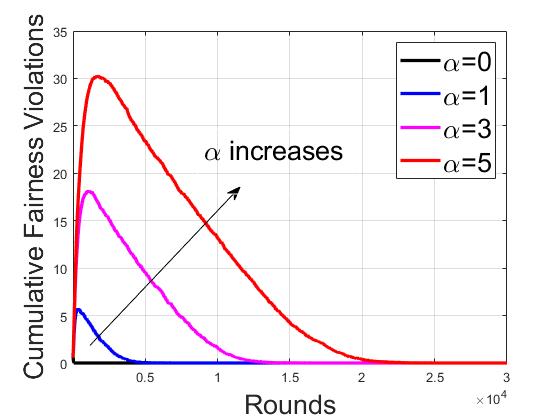}
\hspace{-0.28in}
} 
\subfloat[Short-term Reward Regularity]{
\label{fig:sim:alpha_reward}
\includegraphics[scale=0.24]{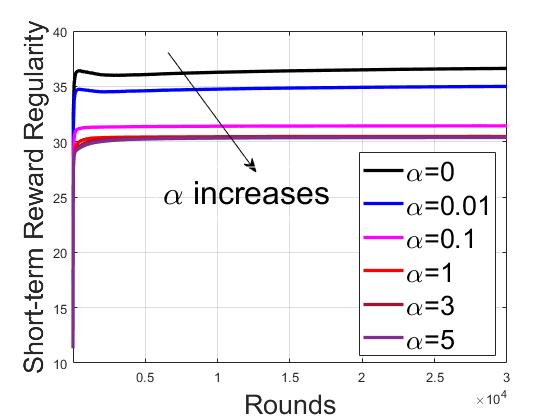}
\hspace{-0.28in}
} 
\subfloat[Cumulative Regret]{ \label{fig:sim:alpha_regret}
\includegraphics[scale=0.24]{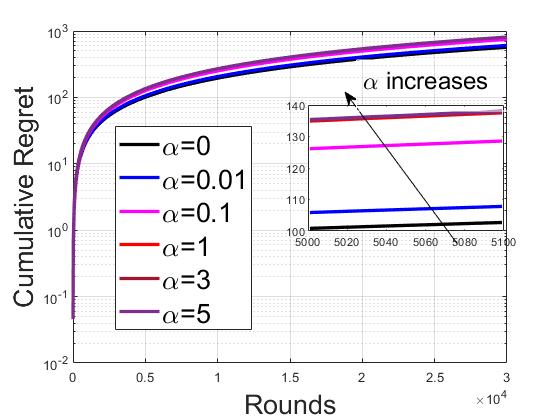}
\hspace{-0.28in}
} 
\caption{Synthetic simulations: impact of parameter $\alpha$.}
\label{fig:sim:alpha}
\end{figure*}

\begin{figure*}[!htbp]
\vspace{-0.3in}
\centering 
\subfloat[Average Reward]{
\label{fig:sim:beta_frac}
\includegraphics[scale=0.24]{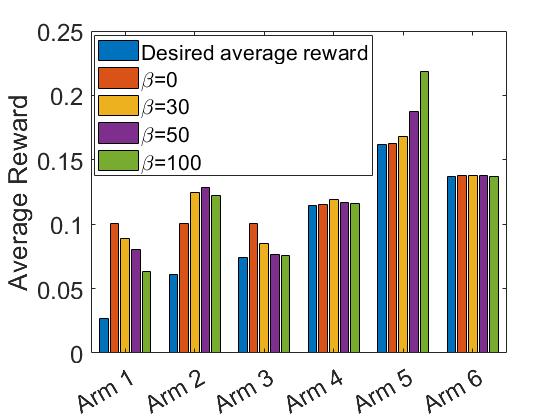}
\hspace{-0.28in}
} 
\subfloat[Cumulative Fairness Violations]{ \label{fig:sim:beta_violation}
\includegraphics[scale=0.24]{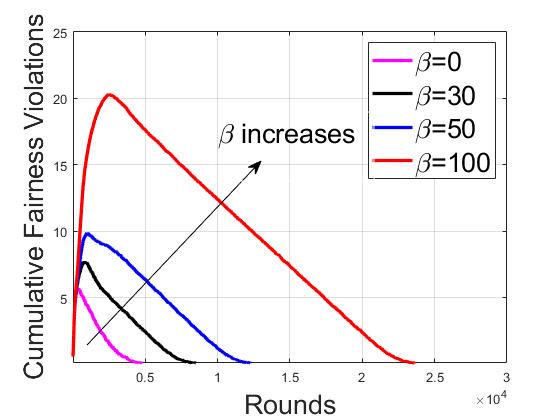}
\hspace{-0.28in}
} 
\subfloat[Short-term Reward Regularity]{
\label{fig:sim:beta_reward}
\includegraphics[scale=0.24]{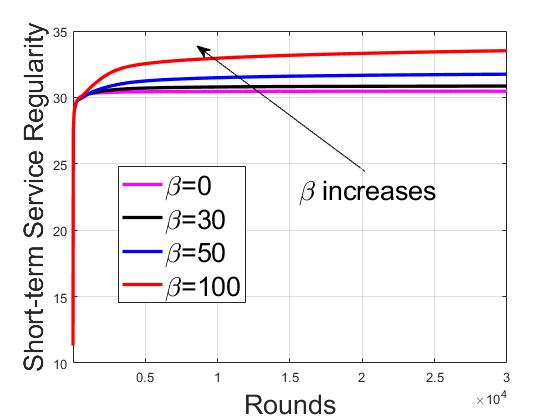}
\hspace{-0.28in}
} 
\subfloat[Cumulative Regret]{ \label{fig:sim:beta_regret}
\includegraphics[scale=0.24]{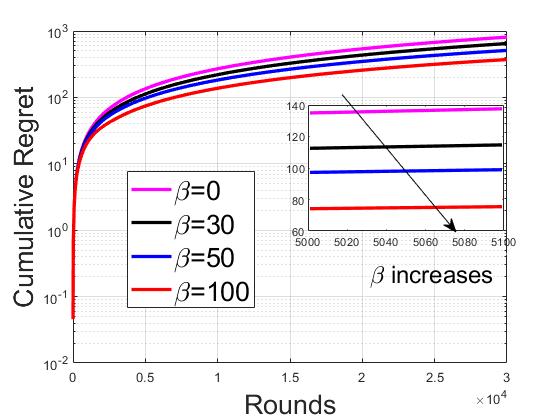}
\hspace{-0.28in}
} 
\caption{Synthetic simulations: impact of parameter $\beta$.}
\label{fig:sim:beta}
\end{figure*}

\begin{figure*}[!htbp]
\vspace{-0.3in}
\centering 
\subfloat[Average Reward]{
\label{fig:sim:trace_alpha_frac}
\includegraphics[scale=0.24]{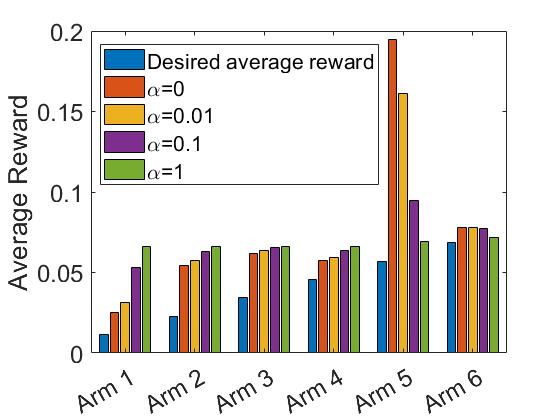}
\hspace{-0.28in}
} 
\subfloat[Cumulative Fairness Violations]{ \label{fig:sim:trace_alpha_violation}
\includegraphics[scale=0.24]{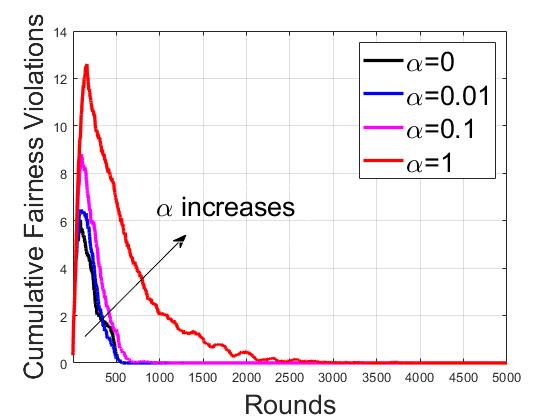}
\hspace{-0.28in}
} 
\subfloat[Short-term Reward Regularity]{
\label{fig:sim:trace_alpha_reward}
\includegraphics[scale=0.24]{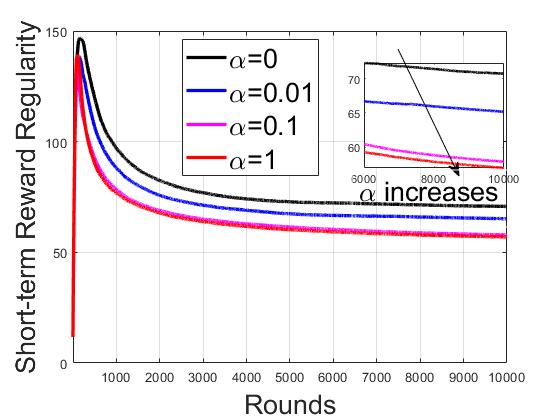}
\hspace{-0.28in}
} 
\subfloat[Cumulative Regret]{ \label{fig:sim:trace_alpha_regret}
\includegraphics[scale=0.24]{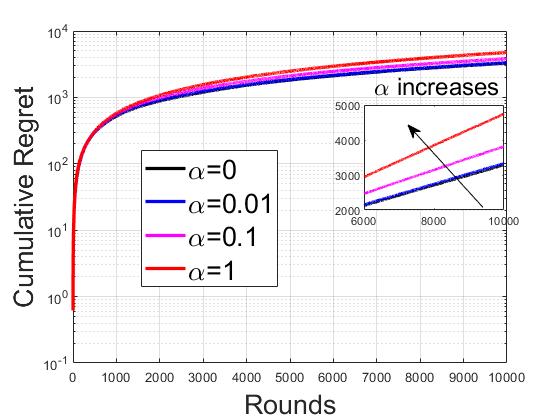}
\hspace{-0.28in}
} 
\caption{Multi-user panoramic scene delivery: impact of parameter $\alpha$.}
\label{fig:sim:trace_alpha}
\end{figure*}

\begin{figure*}[!htbp]
\vspace{-0.3in}
\centering 
\subfloat[Average Reward]{
\label{fig:sim:trace_beta_frac}
\includegraphics[scale=0.24]{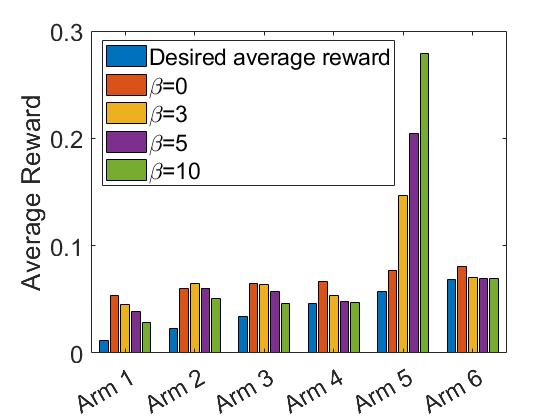}
\hspace{-0.28in}
} 
\subfloat[Cumulative Fairness Violations]{ \label{fig:sim:trace_beta_violation}
\includegraphics[scale=0.24]{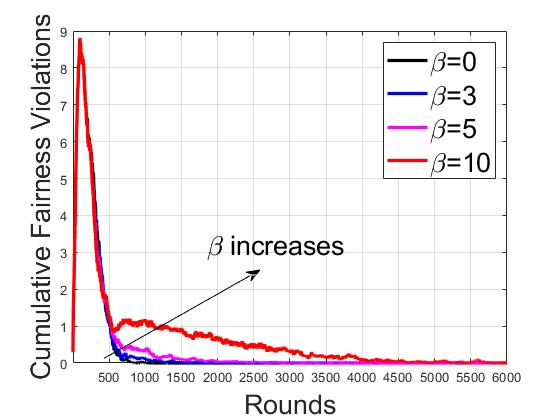}
\hspace{-0.28in}
} 
\subfloat[Short-term Reward Regularity]{
\label{fig:sim:trace_beta_reward}
\includegraphics[scale=0.24]{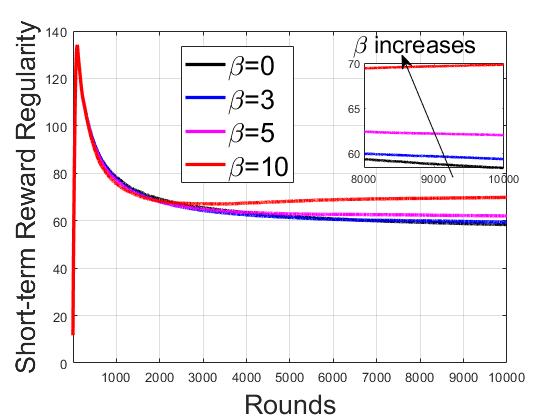}
\hspace{-0.28in}
} 
\subfloat[Cumulative Regret]{ \label{fig:sim:trace_beta_regret}
\includegraphics[scale=0.24]{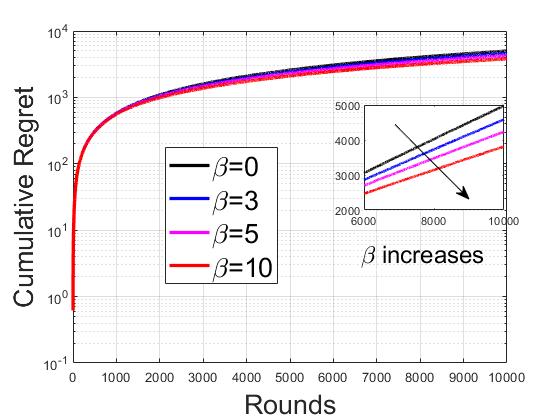}
\hspace{-0.28in}
} 
\caption{Multi-user panoramic scene delivery: impact of parameter $\beta$.}
\label{fig:sim:trace_beta}
\end{figure*}
\vspace{-0.1in}

\begin{figure*}[!htbp]
\vspace{-0.3in}
\centering 
\subfloat[Average Reward]{
\label{fig:sim:trace2_alpha_frac}
\includegraphics[scale=0.24]{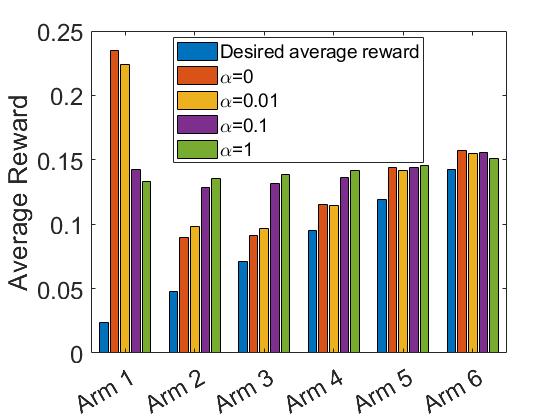}
\hspace{-0.28in}
} 
\subfloat[Cumulative Fairness Violations]{ \label{fig:sim:trace2_alpha_violation}
\includegraphics[scale=0.24]{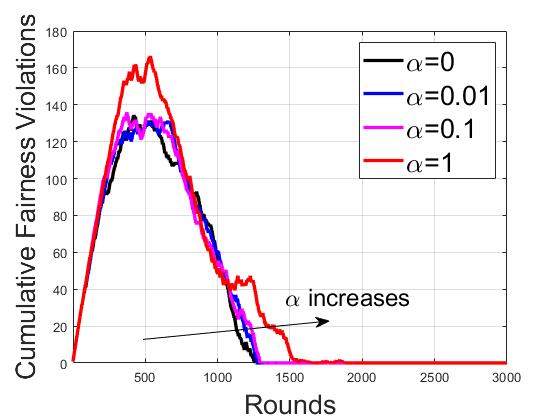}
\hspace{-0.28in}
} 
\subfloat[Short-term Reward Regularity]{
\label{fig:sim:trace2_alpha_reward}
\includegraphics[scale=0.24]{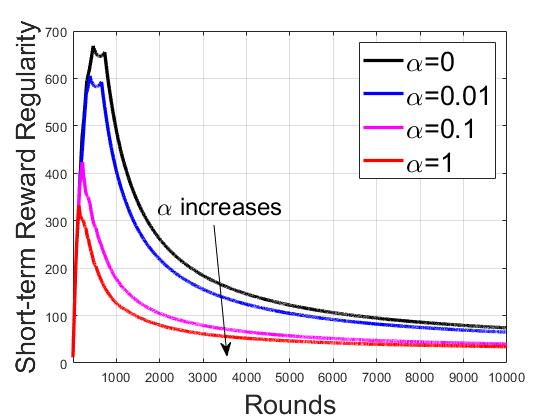}
\hspace{-0.28in}
} 
\subfloat[Cumulative Regret]{ \label{fig:sim:trace2_alpha_regret}
\includegraphics[scale=0.24]{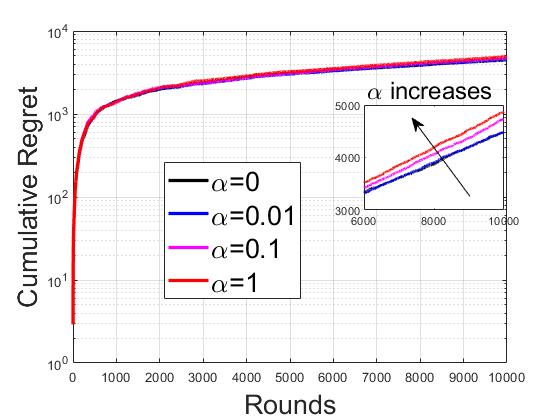}
\hspace{-0.28in}
} 
\caption{Timely information delivery via wireless: impact of parameter $\alpha$.}
\label{fig:sim:trace2_alpha}
\end{figure*}

\begin{figure*}[!htbp]
\vspace{-0.3in}
\centering 
\subfloat[Average Reward]{
\label{fig:sim:trace2_beta_frac}
\includegraphics[scale=0.24]{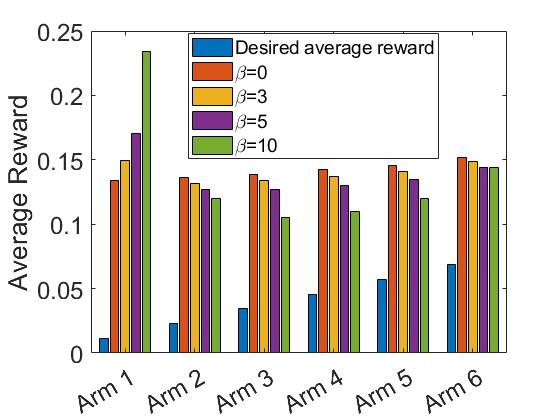}
\hspace{-0.28in}
} 
\subfloat[Cumulative Fairness Violations]{ \label{fig:sim:trace2_beta_violation}
\includegraphics[scale=0.24]{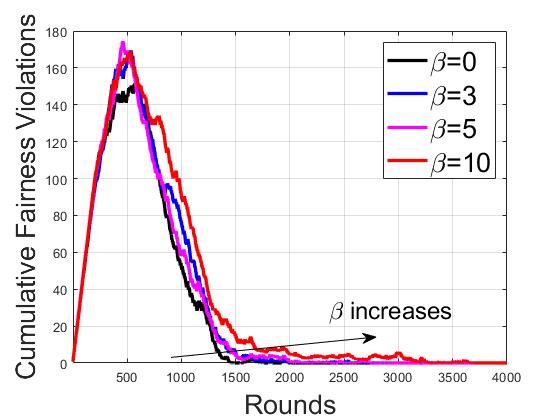}
\hspace{-0.28in}
} 
\subfloat[Short-term Reward Regularity]{
\label{fig:sim:trace2_beta_reward}
\includegraphics[scale=0.24]{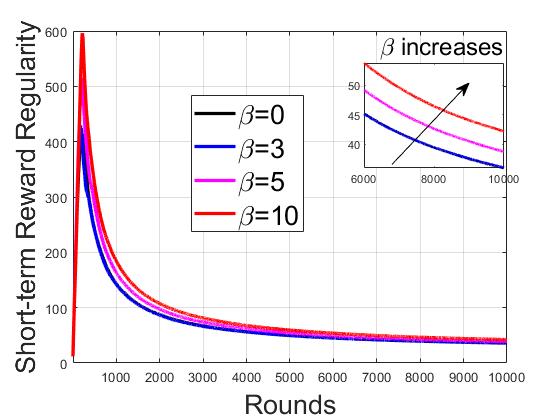}
\hspace{-0.28in}
} 
\subfloat[Cumulative Regret]{ \label{fig:sim:trace2_beta_regret}
\includegraphics[scale=0.24]{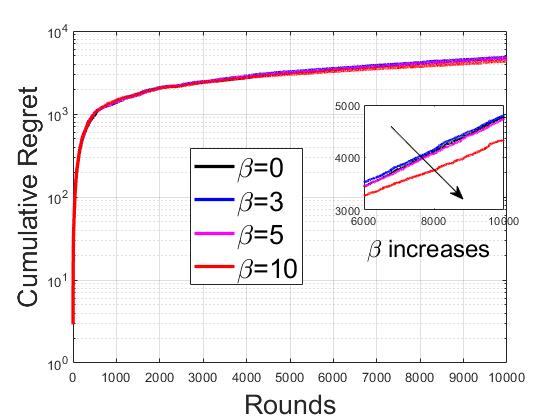}
\hspace{-0.28in}
} 
\caption{Timely information delivery via wireless: impact of parameter $\beta$.}
\label{fig:sim:trace2_beta}
\vspace{-0.2in}
\end{figure*}

\section{Simulations}
In this section, we demonstrate the effectiveness of our proposed RFL algorithm in the applications for multi-user interactive and panoramic scene delivery and timely information delivery via wireless. 

\subsection{Synthetic Simulations}
We consider a timely information delivery application that can be modeled as CMAB formulation with the following setups: the number of arms is $N=6$ (at most one arm can be pulled in each round); the mean reward vector is $\bs{\mu}=(0.7, 0.8, 0.65, 0.75, 0.85, 0.6)$; the minimum required average reward vector is $\bs{\lambda}=0.8\times$(0.7, 1.6, 1.95, 3, 4.25, 3.6)/21; $\epsilon$ is set to $0.001$.

First, we study the impact of parameter $\alpha$ on the performance of cumulative violation, short-term reward regularity, and cumulative regret by fixing parameter $\beta=1$.
Fig. \ref{fig:sim:alpha} shows the performance of RFL algorithm with different $\alpha$ values. We can observe from Fig. \ref{fig:sim:alpha_frac} that the average reward of each arm is larger than its minimum required reward value for each $\alpha$ value, demonstrating that our 
 proposed RFL algorithm achieves long-term fairness. 
In addition, as shown in Fig. \ref{fig:sim:alpha_violation}, when the $\alpha$ value increases, it takes a longer time to achieve zero cumulative fairness violations. This is because a larger $\alpha$ puts more weight on the UCB estimates and thus less weight on the virtual queue-lengths, resulting in achieving zero cumulative fairness violations slower. This also corroborates Proposition \ref{prop:zero_violation}. From Fig. \ref{fig:sim:alpha_reward}, the reward regularity performance improves with the increase of $\alpha$ values, which matches our analytical results (cf.  Proposition \ref{prop:age}) and the intuition that a larger $\alpha$ enforces the RFL algorithm to pull arms with large TSLR values and thus yields better reward regularity performance. However, this is at the cost of degrading cumulative regret performance, as shown in Fig. \ref{fig:sim:alpha_regret}. This is also revealed in Proposition \ref{prop:regret} and matches our intuition. Additionally, we can observe the reward regularity and regret performance do not change evidently after $\alpha$ is larger than 1. This is consistent with our theoretical upper bounds of reward regularity and regret performance that if $\alpha$ is large enough, the reward regularity and the cumulative regret approach some constants.


Next, we investigate the impact of parameter $\beta$ on the system performance by fixing $\alpha=1$. Fig. \ref{fig:sim:beta} shows the performance of our RFL algorithm with varying $\beta$ values. We can observe from Fig. \ref{fig:sim:beta_frac} that the average received reward of each arm is also larger than its minimum required reward value under different $\beta$ values, ensuring long-term fairness. Fig. \ref{fig:sim:beta_violation} shows that our proposed RFL algorithm with a larger $\beta$ value achieves zero cumulative fairness violations slower, which validates Proposition \ref{prop:zero_violation} and matches our intuition, i.e., a larger $\beta$ emphasizes more on the UCB estimates and less on virtual queue-lengths, yielding in poor cumulative fairness violation performance. However, compared with the impact from $\alpha$, parameter $\beta$ has less impact on the cumulative fairness violations. Specifically, $\alpha$ changing from 3 to 5 has a similar effect on the fairness violations as that with $\beta$ varying from 50 to 100, which also matches our theoretical observations. Moreover, as $\beta$ increases, the reward regularity performance deteriorates, as shown in Fig. \ref{fig:sim:beta_reward}, and the cumulative regret performance improves, as shown in Fig. \ref{fig:sim:beta_regret}. These phenomenons validate the correctness of Proposition \ref{prop:age} and Proposition \ref{prop:regret}, i.e., improving reward regularity performance sacrifices the regret performance.

\subsection{Simulations based on Real-World Datasets}
\label{sim:real_world}
\emph{Multi-user Interactive and Panoramic Scene Delivery:}
we demonstrate the effectiveness of our proposed RFL algorithm in a multi-user interactive and panoramic scene delivery application (cf. Section \ref{app:panoramic}) based on real motion trace dataset \cite{bao2016shooting}. This dataset contains motion data from 153 participants who viewed 360-degree videos, capturing three degrees of freedom: pitch, yaw, and roll. In our simulations, we consider $N=6$ users. The AP can at most select one user in one round and send users a panoramic scene. As described in Section \ref{app:panoramic}, the panoramic scene can be partitioned into a finite number of delivery portions. Hence, the AP can decide the portion of the panoramic scene to be delivered to each user in each round. There are $5$ different types of portion, i.e., $(0.625, 0.65, 0.7, 0.75, 1)$. The successful viewing probability of each delivery is the product of the viewport prediction probability and the successful transmission probability. We use the autoregressive model \cite[Algorithm 1]{chen2021motion} to predict the user’s head motion and then use it to calculate the successful viewing probability based on the real head motion trace. In terms of the successful transmission probability, we assume there is $i.i.d.$ ON-OFF channel
fading over time with heterogeneous unknown successful transmission probabilities.
For the fairness constraint, the minimum required average reward vector is $\bs{\lambda}=0.8\times(0.3, 0.6, 0.9, 1.2, 1.5, 1.8)/21$. We set $\epsilon$ to $0.001$.
Fig. \ref{fig:sim:trace_alpha} and Fig. \ref{fig:sim:trace_beta} illustrate the impact of parameters $\alpha$ and $\beta$ on the cumulative fairness, short-term reward regularity, and cumulative regret. The observations are similar to those in Fig. \ref{fig:sim:alpha} and Fig. \ref{fig:sim:beta} via synthetic simulations.

\emph{Timely Information Delivery via Wireless:}
we illustrate the implementation of our proposed RFL algorithm in the application of timely information delivery via wireless (cf. Section \ref{app:rate}) based on a real-world dataset \cite{wu2023joint}. This dataset collects the post
signal-to-noise ratio (post-SNR) of decoded signal constellations from 10 nodes every 1 ms for a total time duration of 20 seconds, where all these nodes are associated with one AP.
In our simulations, we consider $N=6$ nodes. The AP can at most associate with one node in one round and decide the rate of communication with the node. There are nine rates available for selection, i.e., $(0.73, 0.91, 1.46, 1.825, 2.19, 2.37, 2.92, 3.65, 4.38)$ Gbps.
Suppose the AP selects a rate $r$ for a node and the node measures $snr$ (i.e., signal-to-noise ratio) as its postSNR, if $r(snr) \ge r$, the data packet transmission is successful; otherwise, it fails.
To guarantee fairness among nodes, we set each node's desired scheduling fraction as $\bs{\lambda}=0.5\times(1,2,3,4,5,6)/21$. We set $\epsilon$ to $0.001$.
Fig. \ref{fig:sim:trace2_alpha} and Fig. \ref{fig:sim:trace2_beta} depict the effect of parameters $\alpha$ and $\beta$ on the cumulative fairness, short-term reward regularity and cumulative regret. The observations also align with those presented in Fig. \ref{fig:sim:trace_alpha} and Fig. \ref{fig:sim:trace_beta} via multi-user interactive and panoramic scene delivery simulations.

\section{Conclusion and Future Work}
\label{sec:conclusion}

In this paper, we considered the problem of combinatorial multi-armed bandits to minimize cumulative regret over a finite number of rounds while guaranteeing fairness among arms and the short-term reward regularity of each arm. We developed a parameterized maximum-weight type arm-pulling policy. However, it is quite challenging to characterize the performance of our proposed algorithm due to the strong coupling between the virtual queue-lengths and TSLR metrics and the sharp dynamics of TSLR. We addressed these challenges by revealing a key relationship between the virtual queue-lengths and the TSLR metrics and performing Lyapunov drift analysis based on several non-trivial Lyapunov functions, and successfully analyzed the performance of our proposed algorithm. The theoretical findings were further verified by simulations based on two real-world datasets.

While the tradeoff between the reward regularity and the cumulative regret was successfully characterized under our proposed algorithm and is tight in certain cases, as discussed in Remark \ref{remark:tradeoff}, however, it is unclear whether such a tradeoff is indeed optimal, which requires further investigation. Moreover, our proposed algorithm shares the same computational complexity with classical CMAB algorithms, and requires new research efforts on its low-complexity algorithm design and performance tradeoff characterization among all metrics.

\section{acknowledgements}
The authors would like to thank Juaren Steiger at Queen's University for his valuable comments on this paper.

\appendices



\section{Proof of Proposition \ref{prop:zero_violation}}
\label{App:proof:zero_violation}

Select the Lyapunov function
\begin{align}
V(t) \triangleq \|\mb{W}(t)\|_2,
\end{align}
where $\mb{W}(t)\triangleq (\mb{Q}(t)/\sqrt{\bs{\mu}},2\sqrt{\alpha\mb{Z}(t)/\bs{\mu}})$. Let $\mb{I}(t)\triangleq(\mb{Q}(t),\mb{Z}(t),\mb{w}(t))$. Then, we have the following key lemma that characterizes the conditional expected drift when $V(t)$ is large enough, which is shown in Appendix \ref{APP:lemma:drift}.
\begin{lemma}
\label{lemma:drift}
For any $0<\epsilon\leq\delta/2$, if $V(t)\geq U(\alpha,\beta)\triangleq8N^2(4\alpha+3\beta+2)/(\delta\mu_{\min}^2)$, 
then 
\begin{align}
\bE\left[V(t+1)-V(t)\middle|\mb{I}(t)\right]\leq -\frac{\delta \mu_{\min}}{16N}.
\end{align}
Moreover, 
\begin{align*}
\left|V(t+1)-V(t)\right|
\leq \frac{1}{\lambda_{\min}\mu_{\min}}\left(12\alpha+1\right)N\triangleq D(\alpha),
\end{align*}
where we recall that $\lambda_{\min}\triangleq\min_{n}\lambda_n>0$ and $\mu_{\min}\triangleq\min_{n}\mu_{n}>0$.
\end{lemma}

For any $\epsilon\leq\delta/2$, Lemma \ref{lemma:drift} satisfies the conditions of \cite[Lemma 11]{liu2021efficient} and thus we have 
\begin{align}
\bE\left[e^{\theta(\alpha) V(t)}\right]\leq e^{\theta(\alpha)  V(0)}+v_0(\alpha)e^{\theta(\alpha)(D(\alpha)+U(\alpha,\beta))},     
\end{align}
where $\theta(\alpha)=3\delta \mu_{\min}/(48ND^2(\alpha)+ \mu_{\min}\delta D(\alpha))$ and $v_0(\alpha)\triangleq32N/(\delta \mu_{\min}\theta(\alpha))$.

According to Jensen's inequality for convex function $e^{\theta x}$, we have 
\begin{align*}
e^{\theta(\alpha)\bE[V(t)]}&\leq\bE\left[e^{\theta(\alpha) V(t)}\right] \nonumber \\
&\leq e^{\theta(\alpha)  V(0)}+v_0(\alpha)e^{\theta(\alpha)(D(\alpha)+ U(\alpha,\beta))},
\end{align*}
which implies 
\begin{align}
&\bE[V(t)]\leq\frac{1}{\theta(\alpha)}\log\left(e^{\theta(\alpha)  V(0)}+v_0(\alpha) e^{\theta(\alpha)(D(\alpha)+ U(\alpha,\beta))}\right)\nonumber\\
\stackrel{(a)}{\leq}&\frac{1}{\theta(\alpha)}\log\left((v_0(\alpha)+1)e^{\theta(\alpha)(V(0)+D(\alpha)+ U(\alpha,\beta))}\right)\nonumber\\
\stackrel{(b)}{=}&\frac{1}{\theta(\alpha)}\log (v_0(\alpha)+1)+D(\alpha)+ U(\alpha,\beta),
\end{align}
where step $(a)$ is true since $v_0(\alpha)\geq1$; $(b)$ is true for $V(0)=0$.


According to the dynamics of virtual queues (cf. \eqref{eqn:virtualQ}), we have 
\begin{align*}
Q_n(t+1)\geq Q_n(t)+\lambda_n-\wh{S}_n(t)X_n(t)+\epsilon.     
\end{align*}
By summing the above inequality over $\tau=0,1,\ldots,t-1$ and taking expectation on both sides, we have
\begin{align}
\bE[Q_n(t)]\geq\sum_{\tau=0}^{t-1}(\lambda_n-\bE[\wh{S}_n(\tau)X_n(\tau)])+\epsilon t.
\end{align}
Using the fact that $V(t)\geq\|\mb{Q}(t)/\mb{\sqrt{\bs{\mu}}}\|_2\geq\|\mb{Q}(t)\|_1/\sqrt{N}$ due to $\mu_n\leq1, \forall n$, we have 
\begin{align}
\label{eqn:prop:vlo:queue_length}
\bE\left[\|\mb{Q}(t)\|_1\right]\leq\sqrt{N}\left(\frac{1}{\theta(\alpha)}\log (v_0(\alpha)+1)+D(\alpha)+U(\alpha,\beta)\right). 
\end{align}
As such, we have 
\begin{align}
\label{eqn:prop:vol:main}
&\left(\sum_{\tau=0}^{t-1}(\lambda_n-\bE[\wh{S}_n(\tau)X_n(\tau)])\right)^{+}\leq\left(\bE[Q_n(t)]-\epsilon t\right)^{+}\nonumber\\
\leq&\left(\bE[\|\mb{Q}(t)\|_1]- \epsilon t\right)^{+}\nonumber\\
\leq & \left( g_0(\alpha,\beta)-\epsilon t\right)^{+} .
\end{align}
where $g_0(\alpha,\beta)\triangleq\sqrt{N}\bigg(\frac{1}{\theta(\alpha)}\log (v_0(\alpha)+1)+D(\alpha)+U(\alpha,\beta) \bigg)$.

\section{Proof of Proposition \ref{prop:age}}
\label{APP:proof:age}
In this section, we derive the second part of the upper bound on reward regularity performance. Consider the Lyapunov function
\begin{align}
V_1(t) \triangleq \|\mb{W}(t)\|_2^2,
\end{align}
where we recall that $\mb{W}(t)\triangleq (\mb{Q}(t)/\sqrt{\mb{\mu}},2\sqrt{\alpha\mb{Z}(t)/\mb{\mu}})$. In the rest of the proof, we omit the round index $t$ properly without introducing any confusion. We also use $Y^{+}$ to denote $Y(t+1)$ for variable $Y$. Next, we consider the conditional expected drift $\Delta V_1\triangleq\bE\big[V_1^+-V_1|\mb{I}\big]$ given $\mb{I}=(\mb{Q},\mb{Z},\mb{w})$. 
\begin{align}
\label{eqn:prop:age:V1}
&\Delta V_1=\bE\left[\sum_{n=1}^{N}\frac{(Q_n^{+})^2}{\mu_n}+4\alpha\sum_{n=1}^{N}\frac{Z_n^{+}}{\mu_n}-\sum_{n=1}^{N}\frac{Q_n^2}{\mu_n}-4\alpha\sum_{n=1}^{N}\frac{Z_n}{\mu_n}\middle|\mb{I}\right] \nonumber\\
\stackrel{(a)}{\leq}&\sum_{n=1}^{N}\bE\left[\frac{(Q_n+\lambda_n-\wh{S}_nX_n+\epsilon)^2}{\mu_n}-\frac{Q_n^2}{\mu_n}+4\alpha\frac{Z_n^{+}-Z_n}{\mu_n}\middle|\mb{I}\right]\nonumber\\
\stackrel{(b)}{\leq}&\sum_{n=1}^{N}\bE\left[\frac{2}{\mu_n}Q_n (\lambda_n+\epsilon-\wh{S}_nX_n)+\frac{1}{\mu_n}\left(\lambda_n-\wh{S}_nX_n+\epsilon\right)^2\middle|\mb{I}\right]\nonumber\\
&-4\alpha\bE\left[\sum_{n=1}^{N}Z_n\wh{S}_n\middle|\mb{I}\right]+ \frac{4\alpha N}{\mu_{\min}} \nonumber \\
\stackrel{(c)}{\leq} &2\sum_{n=1}^{N}\frac{1}{\mu_n}Q_n (\lambda_n+\epsilon)- 2\sum_{n=1}^{N}\bE\left[\left(Q_n+2\alpha Z_n\right)\wh{S}_n \middle|\mb{I}\right]+ B_1,
\end{align}
where step $(a)$ uses the dynamics of the virtual queue (cf. \eqref{eqn:virtualQ}); $(b)$ is true since 
\begin{align}
\label{eqn:prop:age:dyamics}
\sum_{n=1}^{N}\frac{Z_n^{+}}{\mu_n}=&\sum_{n=1}^{N}\frac{(Z_n+1)(1-\wh{S}_nX_n)+\wh{S}_nX_n}{\mu_n}\nonumber\\
=&\sum_{n=1}^{N}\frac{Z_n}{\mu_n}-\sum_{n=1}^{N}\frac{Z_n\wh{S}_nX_n}{\mu_n}+\sum_{n=1}^{N}\frac{1}{\mu_n},
\end{align}
by following the dynamics of the age (cf. \eqref{eqn:age:dynamics}),  and the reward $X_n$ is independent of the system state and decision; $(c)$ is true for $B_1\triangleq (4\alpha+1) N/\mu_{\min}$ and uses the fact that $\lambda_n+\epsilon\leq\mu_n\leq1$.

Given system state $\mb{I}$, according to the definition of the RFL algorithm, we have 
\begin{align}
\label{eqn:prop:age:RFL}
\sum_{n=1}^{N}\left(Q_n + \alpha Z_n+\beta w_n\right)\wh{S}_n
\geq&\sum_{n=1}^{N}\left(Q_n + \alpha Z_n+\beta w_n\right)S^{\dagger}_n\nonumber\\
\geq&\sum_{n=1}^{N}Q_nS^{\dagger}_n,
\end{align}
where $\mb{S}^{\dagger}\triangleq(S_n^{\dagger})_{n=1}^{N}\in\argmax_{\mb{S}\in\mc{S}}\sum_{n=1}^{N}Q_nS_n$. Hence, we have 
\begin{align}
\label{eqn:prop:age:maxWeight}
\sum_{n=1}^{N}Q_n\wh{S}_n
\geq\sum_{n=1}^{N}Q_nS^{\dagger}_n- \alpha\sum_{n=1}^{N}Z_n\wh{S}_n-\beta\sum_{n=1}^{N}w_n\wh{S}_n.
\end{align}

By substituting \eqref{eqn:prop:age:maxWeight} into \eqref{eqn:prop:age:V1}, we have 
\begin{align}
\label{eqn:prop:age:V2}
\Delta V_1\leq& 2\sum_{n=1}^{N}\frac{1}{\mu_n}Q_n (\lambda_n+\epsilon)-2\sum_{n=1}^{N}Q_nS^{\dagger}_n+B_1\nonumber\\
&+2\beta\sum_{n=1}^{N}w_n\wh{S}_n - 2\alpha\sum_{n=1}^{N}Z_n\wh{S}_n.
\end{align}
Note that there exists non-negative numbers $\zeta(\mb{s})$ satisfying 
\begin{align}
&\lambda_n+\delta\leq\sum_{\mb{s}\in\mc{S}}\zeta(\mb{s})s_n\mu_n, \forall n\\
&\sum_{\mb{s}}\zeta(\mb{s})=1. \label{eqn:prop:age:prob}
\end{align}
Hence, we have 
\begin{align}
\label{eqn:prop:age:capacity}
 \sum_{n=1}^{N}\frac{(\lambda_n+\delta)Q_n}{\mu_n}
\leq&\sum_{\mb{s}\in\mc{S}}\zeta(\mb{s})\sum_{n=1}^{N}Q_ns_n \nonumber\\
\leq&\sum_{\mb{s}\in\mc{S}}\zeta(\mb{s})\sum_{n=1}^{N}Q_nS_n^{\dagger}\nonumber\\
=&\sum_{n=1}^{N}Q_nS_n^{\dagger},
\end{align}
where the second last step follows from the definition of $\mb{S}^{\dagger}$, and the last step uses \eqref{eqn:prop:age:prob}.

By substituting \eqref{eqn:prop:age:capacity} into \eqref{eqn:prop:age:V2} and noting that $\epsilon\leq\delta/2$, and $\mu_n\leq1$, we have 
\begin{align}
\Delta V_1\leq -\delta\sum_{n=1}^{N}Q_n-2\alpha\sum_{n=1}^{N}Z_n\wh{S}_n+B_2,
\end{align}
where $B_2\triangleq B_1 + 2\beta N=(4\alpha+1)N/\mu_{\min}+2\beta N$.

By noting that $\delta\leq1$, we have 
\begin{align}
\Delta V_1 \leq&-\delta\sum_{n=1}^{N}\left(Q_n+ \alpha Z_n+\beta w_n\right)\wh{S}_n+\delta\beta \sum_{n=1}^{N}w_n\wh{S}_n+B_2\nonumber\\
\leq&-\frac{\delta}{N}\sum_{n=1}^{N}\left(Q_n+ \alpha Z_n+\beta w_n\right)+B_3 \label{eqn:prop:age:V0} \\
\leq & -\frac{\delta\alpha}{N} \sum_{n=1}^{N}Z_n +B_3, \label{eqn:prop:age:V}
\end{align}
where the second last step is true for $B_3\triangleq B_2+\beta N=(4\alpha+1)N/\mu_{\min}+3\beta N$, and follows from the fact that $\sum_{n=1}^{N}(Q_n+ \alpha Z_n + \beta w_n)\wh{S}_n\geq\max_{n}(Q_n+ \alpha Z_n+\beta w_n)\geq\sum_{n=1}^{N}(Q_n+ \alpha Z_n+\beta w_n)/N$.

Taking the expectation on both sides of \eqref{eqn:prop:age:V}, we have
\begin{align}
\bE\left[V_1^+ - V_1 \right]
\leq -\frac{\delta\alpha}{N} \sum_{n=1}^{N}\bE[Z_n] + B_3
\end{align}
Summing the above inequality over round $t = 0, 1, \cdots, T-1$,
we have
\begin{align}
\bE\left[V_1(T) - V_1(0) \right]
\leq -\frac{\delta\alpha}{N} \sum_{t=0}^{T-1}\sum_{n=1}^{N}\bE[Z_n(t)] +B_3T
\end{align}
Rearranging the above inequality and utilizing the fact that $V_1(0)=0$, we have 
\begin{align*}
\frac{1}{T}\sum_{t=0}^{T-1}\sum_{n=1}^{N}\bE[Z_n(t)] \leq \frac{B_3N}{\delta\alpha}\leq\frac{N^2}{\delta\mu_{\min}} \left( 1 + \frac{3\beta+4}{\alpha} \right).
\end{align*}

\section{Proof of Proposition \ref{prop:regret}}
\label{App:proof:regret}

We rewrite the regret of the RFL Algorithm as
\begin{align}
\text{Reg}(T)\triangleq& \sum_{t=0}^{T-1}\sum_{n=1}^{N}\left(\bE\left[\mu_nS_n^*(t)\right] -\bE\left[\mu_n\wh{S}_n(t)\right]\right)\nonumber\\
=&\sum_{t=0}^{T-1}\Delta R(t),
\end{align}
where $\Delta R(t)\triangleq \sum_{n=1}^{N}\bE\left[\mu_nS_n^*(t)-\mu_n\wh{S}_n(t)\right]$. 

Next, we consider the drift of the Lyapunov function $L(t)\triangleq\frac{1}{2}\sum_{n=1}^{N}\frac{Q_n^2(t)}{\mu_n}+\alpha \sum_{n=1}^{N}\frac{Z_n(t)}{\mu_n}$. By following similar steps as in \eqref{eqn:prop:age:V1} and adding the term $\beta\Delta R(t)$ on both sides of the drift, we obtain 
\begin{align*}
&\bE\left[L(t+1)-L(t)\right]+\beta \Delta R(t)\nonumber\\
&\stackrel{(a)}{\leq}\sum_{n=1}^{N}\bE\left[\frac{Q_n(t)(\lambda_n-\wh{S}_n(t)X_n(t)+\epsilon)}{\mu_n}\right] \nonumber\\
& - \alpha \sum_{n=1}^{N}\bE\left[\frac{Z_n(t)\wh{S}_n(t)X_n(t)}{\mu_n}\right]\nonumber\\
&+\beta\sum_{n=1}^{N}\bE\left[\mu_nS_n^*(t)\right]-\beta \sum_{n=1}^{N}\bE\left[\mu_n\wh{S}_n(t)\right]+\frac{\left(\alpha +1\right)N}{\mu_{\min}}\nonumber\\
&\stackrel{(b)}{=}\sum_{n=1}^{N}\bE\left[(Q_n(t)+ \alpha Z_n(t)+\beta \mu_n)(S_n^{*}(t)-\wh{S}_n(t))\right]\nonumber\\
&-\alpha \sum_{n=1}^{N}\bE\left[Z_n(t)S_n^*(t)\right]+\sum_{n=1}^{N}\bE\left[Q_n(t)(\frac{\lambda_n+\epsilon}{\mu_n}-S_n^*(t))\right]\nonumber\\
&+\frac{\left(\alpha + 1\right)N}{\mu_{\min}}\nonumber\\
&\stackrel{(c)}{\leq} \frac{\left(\alpha +1 \right)N}{\mu_{\min}} +\sum_{n=1}^{N}\bE\bigg[(Q_n(t)+\alpha Z_n(t)+\beta \mu_n)\nonumber\\
&\qquad\qquad\qquad\qquad\qquad\qquad\qquad\qquad\cdot(S_n^{*}(t)-\wh{S}_n(t))\bigg],
\end{align*}
where step $(a)$ uses \eqref{eqn:prop:age:dyamics} and $\epsilon\leq\delta\leq 1$; $(b)$ uses the fact that $\bE\left[X_n(t)\right]=\mu_n$; $(c)$ follows from the fact that $\mb{S}^*$ is the stationary randomized policy that is independent of the current system state and $\bE[S_n^*(t)X_n(t)]\geq\lambda_n+\epsilon, \forall n$ holds for $\epsilon\leq\delta\leq 1$, and the fact that $\sum_{n=1}^{N}\bE\left[Z_n(t)S_n^*(t)\right]\geq0$.

Dividing $\beta$ on both sides of the above inequality and summing over $t=0, 1,\ldots,T-1$, we have 
\begin{align}
\label{eqn:prop:reg:sumt}
& \text{Reg}(T) = \sum_{t=0}^{T-1}\Delta R(t) \nonumber\\
&\leq -\frac{1}{\beta}\bE\left[L(T)-L(0)\right] + \frac{NT}{\mu_{\min}}\left(\frac{\alpha +1}{\beta}\right) \nonumber\\
&+\sum_{t=0}^{T-1}\frac{1}{\beta}\sum_{n=1}^{N}\bE\left[\left(Q_n(t)+\alpha Z_n(t)+\beta\mu_n\right)\left(S_n^*(t)-\wh{S}_n(t)\right)\right] \nonumber\\
&\leq \frac{NT}{\mu_{\min}}\left(\frac{\alpha + 1}{\beta}\right)+\sum_{t=0}^{T-1}\frac{1}{\beta}\sum_{n=1}^{N}\bE\bigg[\left(Q_n(t)+\alpha Z_n(t)+\beta\mu_n\right)\nonumber \\ 
&\qquad\qquad\qquad\qquad\qquad\qquad\qquad\qquad\cdot\left(S_n^*(t)-\wh{S}_n(t)\right)\bigg].
\end{align}

Next, we focus on the term $$\sum_{n=1}^{N}\left(Q_n(t)+\alpha Z_n(t)+\beta\mu_n\right)\left(S_n^*(t)-\wh{S}_n(t)\right).$$ Then, we have
\begin{align}
\label{eqn:prop:reg:alg}
&\sum_{n=1}^{N}\left(Q_n(t)+ \alpha Z_n(t)+\beta\mu_n\right)\left(S_n^*(t)-\wh{S}_n(t)\right)\nonumber\\
\stackrel{(a)}{\leq}&\sum_{n=1}^{N}\left(Q_n(t)+\alpha Z_n(t)+\beta\mu_n\right)\left(\wt{S}_n(t)-\wh{S}_n(t)\right)\nonumber\\
\stackrel{(b)}{\leq}&\sum_{n=1}^{N}\left(Q_n(t)+\alpha Z_n(t)+\beta\mu_n\right)\left(\wt{S}_n(t)-\wh{S}_n(t)\right)\nonumber\\
&+\sum_{n=1}^{N}\left(Q_n(t)+\alpha Z_n(t)+\beta w_n(t)\right)\left(\wh{S}_n(t)-\wt{S}_n(t)\right)\nonumber\\
\leq &\beta\sum_{n=1}^{N}(w_n(t)-\mu_n)\wh{S}_n(t)+\beta\sum_{n=1}^{N}(\mu_n-w_n(t))\wt{S}_n(t),
\end{align}
where step $(a)$ is true for $$\wt{\mb{S}}(t)\in\argmax_{\mb{S}\in\mc{S}}\sum_{n=1}^{N}\left(Q_n(t)+\alpha Z_n(t)+\beta\mu_n\right)S_n(t),$$
and $(b)$ uses the definition of $\wh{\mb{S}}(t)$.

By substituting \eqref{eqn:prop:reg:alg} into \eqref{eqn:prop:reg:sumt}, we have 
\begin{align}
\label{eqn:prop:reg:final}
\text{Reg}(T) \leq &\frac{NT}{\mu_{\min}}\left(\frac{\alpha + 1}{\beta}\right)+\underbrace{\sum_{t=0}^{T-1}\sum_{n=1}^{N}\bE\left[\left(w_n(t)-\mu_n\right)\wh{S}_n(t)\right]}_{\triangleq G_1(T)} \nonumber\\
& +\underbrace{\sum_{t=0}^{T-1}\sum_{n=1}^{N}\bE\left[\left(\mu_n-w_n(t)\right)\wt{S}_n\right]}_{\triangleq G_2(T)}.
\end{align}


Next, we focus on $G_1(T)$ and $G_2(T)$, respectively. Let $t_{n,\tau}$ denote the round at which arm $n$ successfully received a reward, i.e., $X_n(t_{n,\tau})\wh{S}_n(t_{n,\tau})=1$ and $X_n(t_{n,\tau})\wh{S}_n(t_{n,\tau})=0$ if $t\neq t_{n,\tau}, \tau=1,2,\ldots, H_n(T)$. Therefore, we have $H_n(t_{n,\tau})=\tau-1$.

Let $G_{n,1}(T)\triangleq\sum_{t=t_0}^{T-1}\bE\left[\left(w_n(t)-\mu_n\right)\wh{S}_n(t)\right] $ and thus $G_1(T)=\sum_{n=1}^{N}G_{n,1}(T)$.

Hence, we have 
\begin{align}
\label{eqn:prop:reg:R1}
G_{n,1}(T)\stackrel{(a)}{\leq}&\sum_{t=0}^{T-1}\bE\left[(w_n(t)-\mu_n)\wh{S}_n(t)\id_{\mc{F}_n(t)}\right] \nonumber\\
\stackrel{(b)}{\leq}&\bE\left[\sum_{\tau=1}^{H_n(T)}(w_n(t_{n,\tau})-\mu_n)\id_{\mc{F}_n(t_{n,\tau})}\right] \nonumber\\
\stackrel{(c)}{\leq}&1+\bE\left[\sum_{\tau=2}^{H_n(T)}(w_n(t_{n,\tau})-\mu_n)\id_{\mc{F}_n(t_{n,\tau})}\right]\nonumber\\
\stackrel{(d)}{\leq}&1+\bE\left[\sum_{\tau=2}^{H_n(T)}(w_n(t_{n,\tau})-\mu_n)\id_{\mc{F}_n(t_{n,\tau})\cap\mc{G}_n(t_{n,\tau})}\right] \nonumber \\
& +\sum_{\tau=2}^{\infty}\bE\left[\id_{\ol{\mc{G}}_{n(t_{n,\tau})}}\right],
\end{align}
where step $(a)$ is true for $\mc{F}_n(t)\triangleq\{w_n(t)\geq\mu_n\}$ and $\id_{\{\cdot\}}$ being an indicator function; $(b)$ uses the definition of $t_{n,\tau}$, and the fact that $\wh{S}_n(t)\leq1,\forall t\geq0$; $(c)$ follows from the fact that $w_n(t)\leq1, \forall t\geq0$; $(d)$ is true for 
$$\mc{G}_n(t)\triangleq\left\{\ol{\mu}_n(t)-\mu_n\leq\sqrt{\frac{3\log t}{2H_n(t)}}\right\},$$
and $\ol{\mc{G}}_n(t)$ being the complement of the event $\mc{G}_n(t)$.

Next, we consider the second term on the right hand side (RHS) of \eqref{eqn:prop:reg:R1}. 
\begin{align}
\label{eqn:prop:reg:R1:second}
&\bE\left[\sum_{\tau=2}^{H_n(T)}(w_n(t_{n,\tau})-\mu_n)\id_{\mc{F}_n(t_{n,\tau})\cap\mc{G}_n(t_{n,\tau})}\right] \nonumber\\
\stackrel{(a)}{\leq}&\bE\left[\sum_{\tau=2}^{H_n(T)}2\sqrt{\frac{3\log t_{n,\tau}}{2H_n(t_{n,\tau})}}\right] \nonumber\\
\stackrel{(b)}{\leq}&\sqrt{6\log T}\bE\left[\sum_{\tau=2}^{H_n(T)}\frac{1}{\sqrt{\tau-1}}\right]\nonumber\\
\leq&\sqrt{6\log T}\left(1 + \int_{1}^{H_n(T)}\frac{1}{\sqrt{x}}dx\right)\nonumber\\
\leq&2\sqrt{6\log T}\bE\left[\sqrt{H_n(T)}\right],
\end{align}
where step $(a)$ uses the definition of $w_n(t)$ and $\mc{G}_n(t)$, and $(b)$ follows from the fact that $t_{n,\tau}\leq T$ and the definition of $t_{n,\tau}$. With regard to the third term on the RHS of \eqref{eqn:prop:reg:R1}, we have 
\begin{align*}
&\bE\left[\id_{\ol{\mc{G}}_n(t_{n,\tau})}\right]=\Pr\{\ol{\mc{G}}_n(t_{n,\tau})\}\nonumber\\
\stackrel{(a)}{\leq}&\Pr\left\{\bigcup_{m=\tau-1}^{T-1}\left\{\ol{\mu}_n(m)-\mu_n>\sqrt{\frac{3\log m}{2(\tau-1)}}\right\}\right\}\nonumber\\
\leq&\Pr\left\{\bigcup_{m=\tau-1}^{T-1}\left\{\ol{\mu}_n(m)-\mu_n>\sqrt{\frac{3\log m}{2m}}\right\}\right\}\nonumber\\
\stackrel{(b)}{\leq}&\sum_{m=\tau-1}^{T-1}\Pr\left\{\ol{\mu}_n(m)-\mu_n>\sqrt{\frac{3\log m}{2m}}\right\} \nonumber\\
\stackrel{(c)}{\leq}&\sum_{m=\tau-1}^{T-1}\frac{1}{m^3}\leq\frac{1}{(\tau-1)^3}+\int_{\tau-1}^{\infty}\frac{1}{x^3}dx\stackrel{(d)}{\leq}\frac{3}{2(\tau-1)^2},
\end{align*}
where step $(a)$ follows from the fact that 
$$\ol{\mc{G}}_n(t_{n,\tau})\subset\bigcup_{m=\tau-1}^{T-1}\left\{\ol{\mu}_n(m)-\mu_n>\sqrt{\frac{3\log m}{2(\tau-1)}}\right\};$$
$(b)$ uses the union bound; $(c)$ follows from the Chernoff-Hoeffding Bound (see, e.g., \cite[Fact 1]{auer2002finite}), i.e., for $X_1,X_2,\ldots,X_n$ be i.i.d. random variables with common range $[0,1]$ and mean $\mu$, then for any $a\geq0$, we have 
\begin{align}
\label{eqn:prop:reg:chernoff}
\Pr\left\{\frac{1}{n}\sum_{i=1}^{n}X_i\geq \mu+a\right\}\leq e^{-2na^2},
\end{align}
$(d)$ is true for $\tau\geq2$.

Hence, the third term on the RHS of \eqref{eqn:prop:reg:R1} can be bounded as follows.
\begin{align}
\label{eqn:prop:reg:R1:third}
\sum_{\tau=2}^{\infty}\bE\left[\id_{\ol{\mc{G}}_n(t_{n,\tau})}\right]
\leq\sum_{\tau=1}^{\infty}\frac{3}{2\tau^2}=\frac{\pi^2}{4},
\end{align}
where the last step use the fact that $\sum_{n=1}^{\infty}1/n^2=\pi^2/6$. By substituting \eqref{eqn:prop:reg:R1:second} and \eqref{eqn:prop:reg:R1:third} into \eqref{eqn:prop:reg:R1} and using the definition of $G_1(T)$, we have 
\begin{align}
\label{eqn:prop:reg:R1:final}
&G_1(T)\leq N\left(1+\frac{\pi^2}{4}\right) + 2\sqrt{6\log T}\sum_{n=1}^{N}\bE\left[\sqrt{H_n(T)}\right]\nonumber\\
\stackrel{(a)}{\leq}&N\left(1+\frac{\pi^2}{4}\right) + 2N\sqrt{6\log T}\bE\left[\sqrt{\frac{1}{N}\sum_{n=1}^{N}H_n(T)}\right] \nonumber\\
\stackrel{(b)}{\leq}&N\left(1+\frac{\pi^2}{4}\right) + 2\sqrt{6N S_{\max}T\log T},
\end{align}
where step $(a)$ uses Jensen's inequality; $(b)$ follows the fact that $\sum_{n=1}^{N}H_n(T)\leq TS_{\max}$.

Next, we consider the term $G_2(T)$. First, we note that  
\begin{align}
G_2(T)\leq\sum_{t=0}^{T-1}\sum_{n=1}^{N}\bE\left[(\mu_n-w_n(t))\wt{S}_n(t)\id_{\ol{\mc{F}}_{n}(t)}\right],
\end{align}
where we recall that $\mc{F}_n(t)\triangleq\{w_n(t) \geq \mu_n\}$. Note that for $t\leq t_{n,1}$, we have $w_n(t)=1$ and thus $\mc{F}_n(t)$ happens. Therefore, we have 
\begin{align}
\label{eqn:prop:reg:R2:final}
&G_2(T)\leq\sum_{n=1}^{N}\bE\left[\sum_{t=t_{n,1}+1}^{T-1}\left(\mu_n-w_n(t)\right)\wt{S}_n(t)\id_{\ol{\mc{F}}_n(t)}\right]\nonumber\\
\stackrel{(a)}{\leq}&\sum_{n=1}^{N}\bE\left[\sum_{t=t_{n,1}+1}^{T-1}\Pr\left\{\ol{\mu}_n(t)-\mu_n\leq-\sqrt{\frac{3\log t}{2H_n(t-1)}}\right\}\right]\nonumber\\
\leq&\sum_{n=1}^{N}\sum_{\tau=1}^{T-1}\sum_{m=1}^{\tau}\Pr\left\{\frac{1}{m}\sum_{i=1}^{m}X(i)-\mu_n\leq-\sqrt{\frac{3\log \tau}{2m}}\right\}\nonumber\\
\stackrel{(b)}{\leq}&\sum_{n=1}^{N}\sum_{\tau=1}^{T-1}\sum_{m=1}^{\tau}\frac{1}{\tau^3}=\sum_{n=1}^{N}\sum_{\tau=1}^{T-1}\frac{1}{\tau^2}\stackrel{(c)}{\leq}\frac{N\pi^2}{6},
\end{align}
where step $(a)$ follows from the fact that $\mu_n\leq 1$ and $\wt{S}_n(t)\leq 1$ as well as the definition of $\ol{\mc{F}}_n(t)$; $(b)$ again uses the Chernoff-Hoeffding Bound (cf. \eqref{eqn:prop:reg:chernoff}); $(c)$ is true since $\sum_{\tau=1}^{T-1}1/\tau^2\leq\sum_{\tau=1}^{\infty}1/\tau^2=\pi^2/6$. 

Hence, by substituting \eqref{eqn:prop:reg:R1:final} and \eqref{eqn:prop:reg:R2:final} into \eqref{eqn:prop:reg:final}, we have the desired result.

\section{Proof of Lemma \ref{lemma:drift}}
\label{APP:lemma:drift}
We consider the conditional expected drift of the Lyapunov function $V(t)=\|\mb{W}(t)\|_2$ given the current state $\mb{I}(t)=(\mb{Q}(t),\mb{Z}(t),\mb{w}(t))$. In the rest of the proof, we omit the round index $t$ properly without causing confusion and use $Y^{+}$ to denote $Y(t+1)$.
\begin{align}
\label{eqn:prop:age:main}
\Delta V\triangleq&\bE\left[V^+-V\middle|\mb{I}\right]\nonumber\\
=&\bE\left[\sqrt{\|\mb{W}^+\|_2^2}-\sqrt{\|\mb{W}\|_2^2}\middle|\mb{I}\right]\nonumber\\
\leq&\frac{1}{2\|\mb{W}\|_2}\bE\left[\|\mb{W}^+\|_2^2-\|\mb{W}\|_2^2\middle|\mb{I}\right]  \nonumber\\
=&\frac{\Delta V_1}{2\|\mb{W}\|_2},
\end{align}
where the second last step uses the fact that $f(x)=\sqrt{x}$ is concave for $x>0$ and thus $f(x_1)-f(x_2)\leq f'(x_2)(x_1-x_2)=(x_1-x_2)/(2\sqrt{x_2})$ with $x_1=\|\mb{W}^{+}\|_2^2$ and $x_2=\|\mb{W}\|_2^2$. 


From inequality \eqref{eqn:prop:age:V0}, we have
\begin{align}
\label{eqn:prop:age:V3}
\Delta V_1 
\stackrel{(a)}{\leq}&-\frac{\delta \mu_{\min} }{4N}\sum_{n=1}^{N}\left(\frac{Q_n}{\mu_n}+ \frac{4\alpha Z_n}{\mu_n}\right)+B_3\nonumber\\
\stackrel{(b)}{\leq}&-\frac{\delta \mu_{\min}}{4N}\sum_{n=1}^{N}\left(\frac{Q_n}{\sqrt{\mu_n}}+ \sqrt{\frac{4\alpha Z_n}{\mu_n}}\right)+B_4\nonumber\\
\stackrel{(c)}{\leq}&-\frac{\delta \mu_{\min}}{4N}\|\mb{W}\|_2+B_4,
\end{align}
where step $(a)$ is true since we recall that $B_3\triangleq (4\alpha+1)N/\mu_{\min} + 3\beta N$ and $\frac{\mu_{\min}}{4\mu_n}\leq\frac{\mu_{\min}}{\mu_n}\leq 1,\forall n$; 
$(b)$ is true for $B_4\triangleq (4\alpha+3\beta+2)N/\mu_{\min}$ and follows from the inequality that $x \geq \sqrt{x}-1, \forall x\geq0$ and $\mu_n \in (0,1]$; $(c)$ uses the fact that $\|\mb{W}\|_1\triangleq\sum_{n=1}^{N}\left(Q_n/\sqrt{\mu_n}+2\sqrt{\alpha Z_n/\mu_n}\right)$ and $\|\mb{x}\|_1\geq\|\mb{x}\|_2$ for any vector $\mb{x}$.



By substituting \eqref{eqn:prop:age:V3} into \eqref{eqn:prop:age:main}, we have 
\begin{align*}
\Delta V\leq \frac{1}{2\|\mb{W}\|_2}\left(-\frac{\delta \mu_{\min}}{4N}\|\mb{W}\|_2+B_4\right)
=-\frac{\delta \mu_{\min}}{8N}+\frac{B_4}{2V},
\end{align*}
where the last step uses the fact that the Lyapunov function $V=\|\mb{W}\|_2$. This
implies that if $\epsilon\leq\delta/2$, whenever $V\geq U(\alpha,\beta)\triangleq 8NB_4/(\delta \mu_{\min})$, then $\Delta V\leq -\delta \mu_{\min}/(16N)$.

In order to develop an upper bound on the absolute Lyapunov drift $|V^+-V|$, we first establish the following lemma, as shown in Appendix \ref{APP:lemma:drift:bound}.

\begin{lemma}
\label{lemma:drift:bound}
If $V(t)\geq U(\alpha,\beta)$, then 
\begin{align*}
& \left|V(t+1)-V(t)\right| \nonumber \\
 \leq &\frac{1}{\mu_{\min}}\left(N+\frac{4\alpha N}{U(\alpha,\beta)}+\frac{4\alpha \sum_{n=1}^{N}Z_n(t)\wh{S}_n(t)X_n(t)}{V(t)}\right).
\end{align*}
\end{lemma}

Next, we will bound the term $\sum_{n=1}^{N}Z_n(t)\wh{S}_n(t)X_n(t)/V(t)$ in Lemmma \ref{lemma:drift:bound}. By using Lemma \ref{eqn:lemma:QT:relation} that captures the relationship between the virtual queue length and the TSLR, we have
\begin{align}
\label{eqn:eqn:prop:age:ageub}
&\frac{1}{V}\sum_{n=1}^{N}Z_n(t)\wh{S}_n(t)X_n(t)\nonumber\\
\leq&\frac{1}{V}\sum_{n=1}^{N}\frac{1}{\lambda_n}\left(1+Q_n(t)\right)\wh{S}_n(t)X_n(t)\nonumber\\
\stackrel{(a)}{\leq}&\frac{N+\|\mb{Q}(t)/\sqrt{\mb{\mu}}\|_1}{\lambda_{\min}V}\nonumber\\
\stackrel{(b)}{\leq}&\frac{N}{\lambda_{\min}U(\alpha,\beta)}+\frac{\sqrt{N}\|\mb{Q}(t)/\sqrt{\mu}\|_2}{\lambda_{\min}V}\nonumber\\
\stackrel{(c)}{\leq}&\frac{N}{\lambda_{\min}U(\alpha,\beta)}+\frac{\sqrt{N}}{\lambda_{\min}}\nonumber\\
\stackrel{(d)}{\leq}&\frac{2N}{\lambda_{\min}}
\end{align}
where step $(a)$ is true for $\lambda_{\min}\triangleq\min_n\lambda_n>0$ and uses the fact that $X_n(t)\leq1$ and $\wh{S}_n(t)\leq1, \forall n$; $(b)$ is true for $V(t)\geq U(\alpha,\beta)$ and follows from the fact that $\|\mb{x}\|_1\leq\sqrt{N}\|\mb{x}\|_2$ for any $N-$dimensional vector $\mb{x}$; $(c)$ uses the fact that $V\geq\|\mb{Q}/\sqrt{\bs{\mu}}\|_2$; $(d)$ is true since $U(\alpha,\beta)\geq1$.

Combining Lemma \ref{lemma:drift:bound} and \eqref{eqn:eqn:prop:age:ageub}, we have 
\begin{align*}
\left|V(t+1)-V(t)\right|
\leq \frac{1}{\lambda_{\min}\mu_{\min}}\left(12\alpha+1\right)N\triangleq D(\alpha).    
\end{align*}

\section{Proof of Lemma \ref{lemma:drift:bound}}
\label{APP:lemma:drift:bound}
If $V\geq U(\alpha,\beta)$, then 
\begin{align}
\label{eqn:lemma:drift:bound:main}
&\left|V^{+}-V\right|
=\left|\|\mb{W}^{+}\|_2-\|\mb{W}\|_2\right| \nonumber\\
\stackrel{(a)}{=}&\frac{\left|\|\mb{W}^{+}\|_2^2-\|\mb{W}\|_2^2\right|}{\|\mb{W}^{+}\|_2+\|\mb{W}\|_2}\nonumber\\
\stackrel{(b)}{\leq}&\frac{\left|\sum_{n=1}^{N}\frac{(Q_n^{+})^2}{\mu_n}-\sum_{n=1}^{N}\frac{Q_n^2}{\mu_n}\right|}{\|\mb{W}^{+}\|_2+\|\mb{W}\|_2}+\frac{4\alpha \left|\sum_{n=1}^{N}\frac{Z_n^{+}}{\mu_n}-\sum_{n=1}^{N}\frac{Z_n}{\mu_n}\right|}{\|\mb{W}^{+}\|_2+\|\mb{W}\|_2}\nonumber\\
\stackrel{(c)}{\leq}&\frac{1}{\mu_{\min}}\left(\left|\|\mb{Q}^{+}\|_2-\|\mb{Q}\|_2\right| +\frac{4\alpha N}{V}+ \frac{4\alpha \sum_{n=1}^{N}Z_n\wh{S}_nX_n }{V}\right)\nonumber\\
\stackrel{(d)}{\leq}&\frac{1}{\mu_{\min}}\left(\left|\|\mb{Q}^{+}\|_2-\|\mb{Q}\|_2\right| +\frac{4\alpha N}{U(\alpha,\beta)}+\frac{4\alpha \sum_{n=1}^{N}Z_n\wh{S}_nX_n}{V}\right),
\end{align}
where step $(a)$ uses the fact that $|x-y|=|\frac{x^2-y^2}{x+y}|=\frac{|x^2-y^2|}{x+y}$ for $x,y>0$; $(b)$ follows from the fact that $|x+y|\leq |x|+|y|$ with $x=\sum_{n=1}^{N}(Q_n^+)^2/\mu_n-\sum_{n=1}^{N}Q_n^2/\mu_n$ and $y=4\alpha(\sum_{n=1}^{N}Z_n^+/\mu_n-\sum_{n=1}^{N}Z_n/\mu_n)$; $(c)$ uses $\eqref{eqn:prop:age:dyamics}$ and we recall that $\mu_{\min}\triangleq\min_{n}\mu_{n}>0$; $(d)$ follows from the condition that $V\geq U(\alpha,\beta)$.

Note that 
\begin{align}
\left|\|\mb{Q}^+\|_2-\|\mb{Q}\|_2\right|
\stackrel{(a)}{\leq}&\|\mb{Q}^+-\mb{Q}\|_2 \nonumber\\
\stackrel{(b)}{\leq}&\|\mb{Q}^+-\mb{Q}\|_1 \nonumber\\
\leq& N\max_n\|Q_n^+-Q_n\|\stackrel{(c)}{\leq} N,
\end{align}
where step $(a)$ uses the fact that $|\|\mb{x}\|_{2}-\|\mb{y}\|_{2}|\leq \|\mb{x}-\mb{y}\|_2$ for vectors $\mb{x}$ and $\mb{y}$; $(b)$ is true since $\|\mb{x}\|_2\leq\|\mb{x}\|_1$; $(c)$ is true since both increment and decrement of a virtual queue is at most $1$. By substituting above inequality into \eqref{eqn:lemma:drift:bound:main}, we have the desired result.

\bibliographystyle{IEEEtran}
\bibliography{refs}

\end{document}